\documentclass[pdflatex,sn-mathphys-num, iicol]{sn-jnl}

\usepackage{graphicx}%
\usepackage{multirow}%
\usepackage{amsmath,amssymb,amsfonts}%
\usepackage{amsthm}%
\usepackage{mathrsfs}%
\usepackage[title]{appendix}%
\usepackage{xcolor}%
\usepackage{textcomp}%
\usepackage{manyfoot}%
\usepackage{booktabs}%
\usepackage{algorithm}%
\usepackage{listings}
\usepackage{algorithmicx}%
\usepackage{algpseudocode}%
\usepackage{listings}%
\usepackage{verbatim}
\usepackage{multirow}
\usepackage{caption}
\usepackage{tikz}
\usepackage{pgfplots}
\usepackage[table,xcdraw]{xcolor}    
\definecolor{mycolor_red}{RGB}{192, 0, 0}
\definecolor{mycolor_green}{RGB}{59, 95, 33}
\definecolor{mycolor_green2}{RGB}{117, 189, 66}
\definecolor{DarkBlue}{RGB}{0,0,128}
\definecolor{ngreen}{RGB}{17, 173, 30}

\theoremstyle{thmstyleone}%
\theoremstyle{thmstyletwo}%

\theoremstyle{thmstylethree}%

\begin{document}

\title[Article Title]{Looking beyond Visible Cues: Implicit Video Question Answering via Dual-Clue Reasoning}


\author[1,2]{\fnm{Tieyuan} \sur{Chen}}

\author[1]{\fnm{Huabin} \sur{Liu}}

\author[3]{\fnm{Yi} \sur{Wang}}

\author[1]{\fnm{Chaofan} \sur{Gan}}

\author[1]{\fnm{Mingxi} \sur{Lv}}

\author[1]{\fnm{Ziran} \sur{Qin}}

\author[1]{\fnm{Shijie} \sur{Li}}

\author[4]{\fnm{Liquan} \sur{Shen}}

\author[5]{\fnm{Junhui} \sur{Hou}}

\author[2,6]{\fnm{Zheng} \sur{Wang}} 

\author*[1,2]{\fnm{Weiyao} \sur{Lin}}\email{wylin@sjtu.edu.cn}

\affil*[1]{\orgname{Shanghai Jiao Tong University}}
\affil[2]{\orgname{ZhongguanCun Academy}}
\affil[3]{\orgname{Shanghai Artificial Intelligence Laboratory}}
\affil[4]{\orgname{Shanghai University}}
\affil[5]{\orgname{City University of Hong Kong}}
\affil[6]{\orgname{Wuhan University}}


\abstract{Video Question Answering (VideoQA) aims to answer natural language questions based on the given video, with prior work primarily focusing on identifying the duration of relevant segments, referred to as explicit visual evidence.
However, explicit visual evidence is not always directly available, particularly when questions target symbolic meanings or deeper intentions, leading to significant performance degradation.
To fill this gap, we introduce a novel task and dataset, \textbf{I}mplicit \textbf{V}ideo \textbf{Q}uestion \textbf{A}nswering (I-VQA), which focuses on answering questions in scenarios where explicit visual evidence is inaccessible.
Given an implicit question and its corresponding video, I-VQA requires answering based on the contextual visual cues present within the video.
To tackle I-VQA, we propose a novel reasoning framework, IRM (Implicit Reasoning Model), incorporating dual-stream modeling of contextual actions and intent clues as implicit reasoning chains.
IRM comprises the Action-Intent Module (AIM) and the Visual Enhancement Module (VEM). AIM deduces and preserves question-related dual clues by generating clue candidates and performing relation deduction. VEM enhances contextual visual representation by leveraging key contextual clues.
Extensive experiments validate the effectiveness of our IRM in I-VQA tasks, outperforming GPT-4o, OpenAI-o3, and fine-tuned VideoChat2 by 0.76\%, 1.37\%, and 4.87\%, respectively. 
Additionally, IRM performs SOTA on similar implicit advertisement understanding and future prediction in traffic-VQA. Datasets and codes are available in the repo: \url{https://github.com/tychen-SJTU/Implicit-VideoQA}.}




\maketitle

\begin{figure*}[t]
\begin{center}
\includegraphics[width=0.99\textwidth]{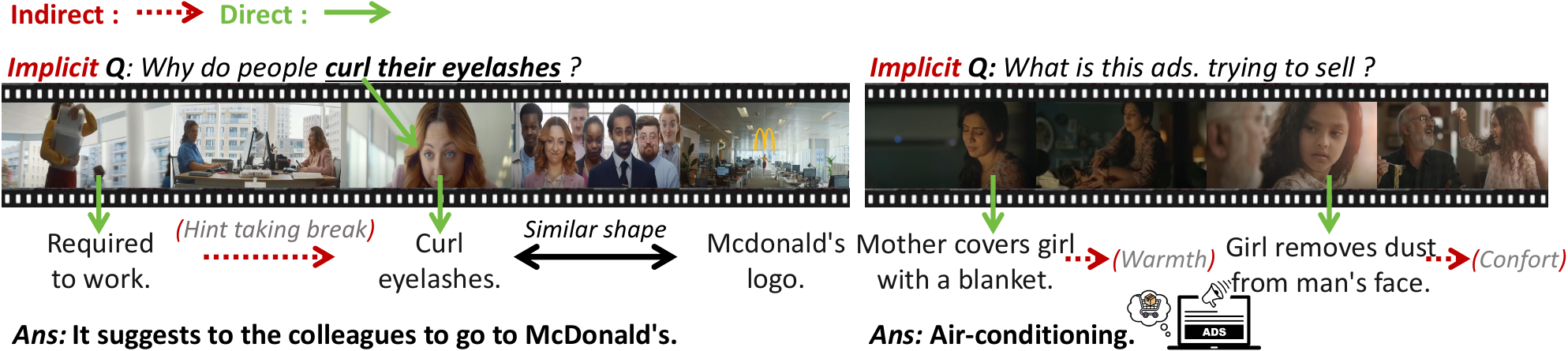}
\end{center}
\caption{\textbf{Examples of Implicit VQA.} Implicit VQA is more challenging due to the non-existence of explicit visual evidence. This happens when questions target symbolic meanings (left figure) or deeper intentions (right figure). Although some information in the question can be directly extracted from the video, and the behavioral action information can be easily captured (\textcolor{mycolor_green2}{as shown by the green arrows}), implicit intentions or inner emotions cannot be simply or directly inferred (\textcolor{mycolor_red}{as shown by the red arrows}).}
\label{fig: intro}
\end{figure*}

\section{Introduction}
\label{sec:intro}
Video Question Answering (VideoQA) is designed to comprehend visual content and generate accurate answers by reasoning over the provided video. 
To facilitate reasoning through more relevant visual segments, an increasing number of VideoQA works have focused on locating key visual segments, referred to as explicit visual evidence. With the explicit evidence, reasoning models (\textit{e.g.}, Large Language Models) can derive more accurate answers~\cite{sevila, gqa1, relation}.
Furthermore, recent works have introduced Grounded Video Question Answering (Grounded-VQA) task~\cite{nextgqa, lita, etbench, rextime, cgbench}, which employs supervised training to locate explicit visual evidence more precisely.

However, the performance of VideoQA is significantly compromised when explicit visual evidence is difficult to locate directly. 
This challenge arises from several factors, including but not limited to ambiguous visual segments~\cite{ambiguous}, perspective shifts~\cite{perspective}, and occlusions~\cite{occlusions}. 
Moreover, grounding explicit evidence becomes particularly challenging for questions that involve symbolic meanings or underlying intentions~\cite{auvid, mmau, vvtut}. 
In these cases, the direct identification of explicit evidence may be confounded by spurious relationships, underscoring the need for a deeper reasoning approach that goes beyond surface-level connections.

As shown in Fig.~\ref{fig: intro}, the first scenario illustrates a case where direct visual grounding faces challenges, as there are no \textbf{direct} reasoning pathways linking the intent of  ``\emph{go to McDonald's}'' to visual segments. 
Instead, through deeper reasoning, we can uncover indirect and implicit relations: the similar shape between the McDonald's logo and the double curled eyelashes.
The second scenario further emphasizes the importance of deeper contextual analysis, where the advertisement's intent is conveyed through implicit environmental cues that evoke warmth and comfort. 

To address these challenges, we set up a new task: \textbf{I}mplict-\textbf{V}ideo \textbf{Q}uestion \textbf{A}nswring (I-VQA), which aims to infer an appropriate answer to implicit video questions, where explicit visual evidence corresponding to the given question cannot be directly accessed or located. 
Meanwhile, we contribute a new dataset for the training and evaluation of I-VQA: We construct and validate implicit video question answering through a semi-automated process, based on existing Grounded-VQA datasets.
However, to our knowledge, no existing solutions can directly tackle implicit questions.

\begin{figure*}[t]
\begin{center}
\includegraphics[width=0.9\textwidth]{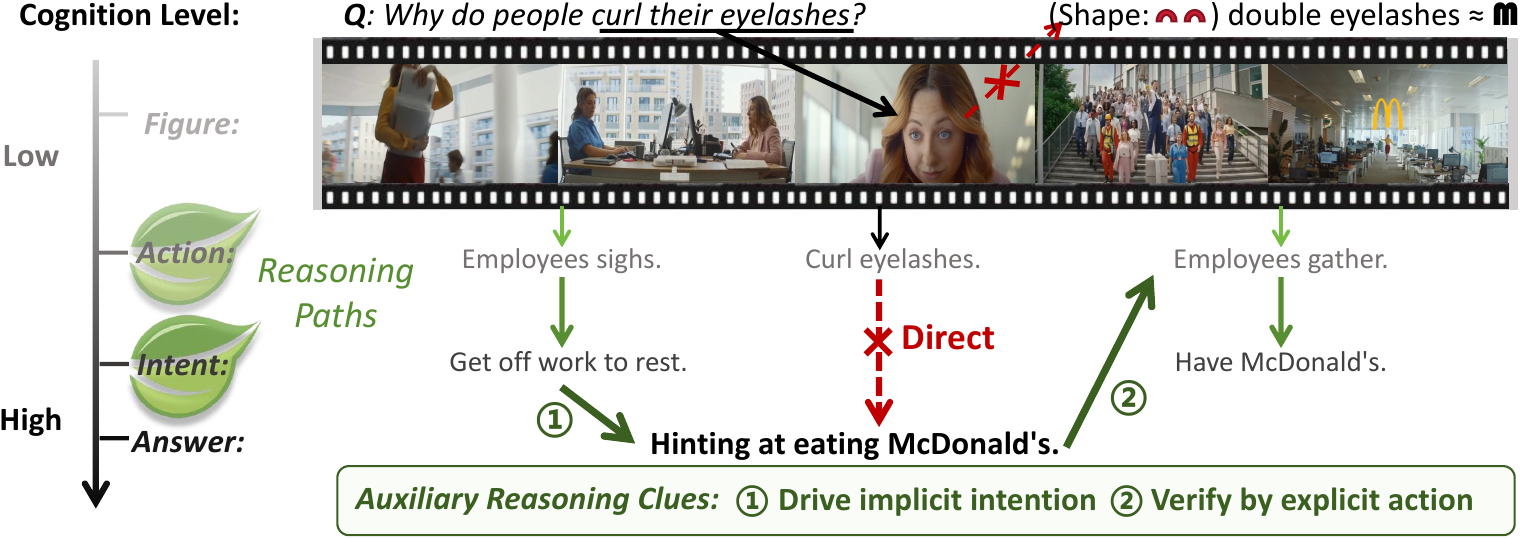}
\end{center}
\caption{\textbf{Dual Clues Aid in Answering Implicit Video Questions.}  When facing implicit questions, direct reasoning fails (\textcolor{mycolor_red}{as indicated by the red arrows}). In contrast, we proposed an effective contextual clue-aided reasoning method (\textcolor{mycolor_green}{as indicated by the green arrows}). Contextual reasoning paths are constructed by first inferring underlying intent from the preceding context, and then verifying through subsequent actions.}
\label{fig: intro_method}
\end{figure*}

To this end, we draw inspiration from works in 
similar evidence-insufficient task: visual commonsense reasoning. These approaches enhance reasoning by leveraging contextual visual information and commonsense to infer additional relevant details~\cite{visualcomet, video2commonsense, visualcommon2, informative, expressive}.
Moreover, in the video modality, richer temporal contextual inputs make it even more crucial to effectively incorporate contextual information.
When further considering the utilized form, cognitive science suggests~\cite{Cognitive} that higher-level information has greater potential for guiding reasoning.
As shown in ~Fig.~\ref{fig: intro_method}, the intention conveyed in the preceding context is often reflected in subsequent actions. 
We can infer implicit actions by extracting intention from the preceding context, while also validating the coherence of these implicit intentions through actions in the following context.
Therefore, we argue that introducing dual clues: (1)~context action involvement clues (2)~context underlying intention clues as additional reasoning pathways is critical for implicit reasoning in the I-VQA task.

Based on this intuition, we introduce IRM, (Implicit Reasoning Model), which employs dual-stream modeling of contextual actions and intent clues as implicit reasoning chains to capture deeper underlying semantics. 
Specifically, we propose the Action-Intent Module (AIM), which deduces textual clue candidates representing explicit contextual actions and their underlying intentions. 
However, these clue candidates may be affected by factors such as hallucination~\cite{MECD, hallucination} and language bias~\cite{biasgpt, biasgpt2}. 
To mitigate error propagation and reduce potential misleading tendencies, we incorporate a visual information verification module within AIM. This module refines dual-clue candidates by verifying them against visual information, leveraging global temporal perception for more accurate validation.

Furthermore, the relationships among sub-events in a video are diverse,  leading to variations in the importance of context clues. Therefore, we propose allocating greater attention to the most relevant context clues. 
Specifically, we introduce a relationship classifier within AIM, which evaluates the relevance of context clues to the implicit question. Only those clues deemed beneficial for implicit question answering are retained, ensuring more effective reasoning.

In addition to constructing textual clues, we introduce the Visual Enhancement Module (VEM) to enhance the visual dual-clue representation. 
VEM applies attention mechanisms between the originally extracted contextual visual information and the refined clues. 
The LLM then performs reasoning with the augmented dual-clue and enhanced visual information. 
Notably, the enhanced visual information can be selectively fed back into the AIM, further mitigating hallucinations and improving relationship deduction. 
This iterative process strengthens the overall modeling of clues.

Evaluation of I-VQA using existing powerful multimodal understanding models, such as GPT-4o, shows that their multi-choice performance is only slightly above 50\%, significantly below human-level performance and the results achieved on other widely-used video datasets. This underscores the considerable room for improvement in implicit video question answering. 
Furthermore, extensive experiments demonstrate the effectiveness and generalizability of IRM for implicit reasoning. IRM achieves state-of-the-art performance not only on the proposed I-VQA dataset but also on current advertisement understanding datasets. 
Additionally, generalization tests highlight IRM’s robust performance in general VideoQA tasks, particularly in predicting outcome events in the SUTD-Traffic dataset and performing general reasoning in MVBench.


\begin{itemize}
\item We introduced the paradigm of implicit video question answering, highlighted the limitations of current models in implicit reasoning capabilities, and emphasized its importance as a largely overlooked yet crucial research direction within VideoQA.
\item We developed the first dataset for implicit video question answering to our knowledge, named I-VQA, through a carefully designed semi-automated pipeline.
\item We demonstrated a robust method for performing implicit reasoning—leveraging dual-stream contextual clue reasoning—and validated its generalizability, showing that it does not compromise the model’s performance on explicit reasoning tasks.
\end{itemize}

\section{Related Work}
\label{sec:related}
\subsection{Video Question Answering.}
Video question answering, as a key representative of video reasoning tasks, has seen rapid development. Recent advancements in VideoQA have been driven by the integration of pre-trained visual encoders with large language models~\cite{videollama, stllm, vqa_1, vqa_2, vqa_3}. 
Notably, several works~\cite{sevila, gqa1, relation} have focused on identifying key visual segments within the temporal domain during the reasoning process, enabling more accurate reasoning.
The Grounded Video Question Answering (Grounded-VQA) task~\cite{nextgqa, lita, etbench, rextime, cgbench, timecraft, vqa_4, video_thought} further strengthens this approach by incorporating human annotations for supervised training.
However, the performance of these methods can significantly degrade when explicit evidence is difficult to locate directly. Therefore, we propose the I-VQA task, which is specifically designed to handle the challenging implicit VideoQA.

\subsection{Visual Context Reasoning.}
Visual context reasoning refers to scenarios where the available information is insufficient for supporting the visual understanding task. Current approaches to visual context reasoning primarily focus on leveraging commonsense knowledge to infer additional visible visual information~\cite{visualcomet, video2commonsense, visualcommon, visualcommon2}. However, these methods are prone to reporting bias and relying on commonsense assumptions can sometimes lead to misguided conclusions.
Another approach models the relationships between contexts, using known context information to infer masked events in video captioning~\cite{VAR} or video causal relation reasoning~\cite{MECD, mecd+}. However, these methods are susceptible to error accumulation, which can negatively affect subsequent reasoning.
We offer a more robust solution to tackle targeted implicit questions; rather than explicitly inferring additional visual data, we construct informative contextual clues consisting of action involvement and semantic intention. 
These clues form reasoning chains that capture deeper underlying semantics, while error accumulation is mitigated through a rigorous refinement process.

\subsection{Implicit Video Reasoning.}
The reasoning tasks related to implicitness in videos mainly focus on inferring implicit attributes of objects, such as mass and charge in virtual videos. 
ComPhy~\cite{comphy} requires models to identify intrinsic properties and make dynamic predictions based on these. 
CRIPP-VQA~\cite{cripp} requires physical properties to be learned from video rather than explicitly expressed in the question. ACQUIRED~\cite{ACQUIRED} is a combination of the two forms mentioned above. 
Another similar task is the PSAV~\cite{adsdataset}, which introduces the persuasion strategy identification task in advertisement understanding, the model needs to distinguish the implicit strategy of the advertisement video among the strategies defined by~\cite{Persuasion}.
However, by constructing a new dataset, our I-VQA focuses on more general implicit questions regarding daily-life videos. 
Moreover, our I-VQA eliminates the possibility of taking explicit evidence as shortcuts, ensuring the rigor of implicit reasoning. 
Therefore, it enables a more rigorous and comprehensive evaluation of the model’s implicit reasoning capability.

\section{I-VQA Task}
\label{sec:task}

\begin{figure*}[!t]
\begin{center}
\includegraphics[width=0.98\textwidth]{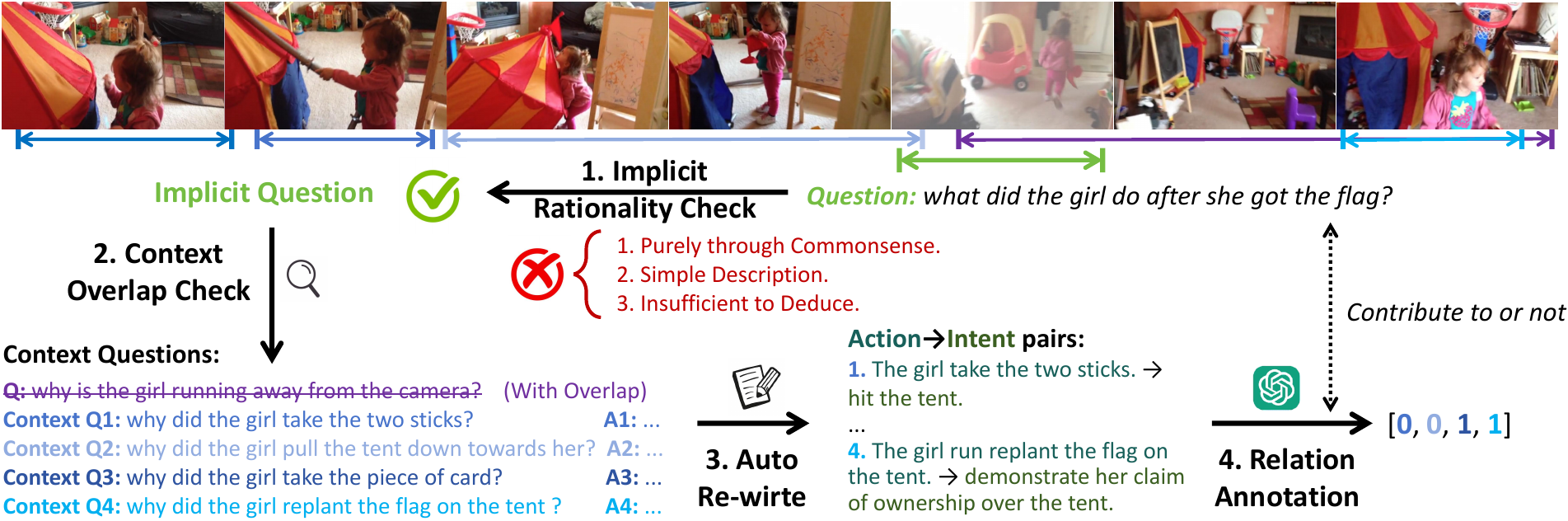}
\end{center}
\caption{\textbf{Dataset Cleaning and Annotation Pipeline.} We carefully check the rigor of the question's implicitness through a semi-automated pipeline, and filter context QA pairs with no overlap to mitigate potential information leakage. We further rewrite context QA pairs as action-intent clues and label the relations between acquiring clues and answering the question.}
\label{fig: anno}
\end{figure*}

In Sec.~\ref{sec: definition}, we introduce our I-VQA task setting, while in Sec.~\ref{sec: dataset} we present the data sources, the principles of collection and annotation, and the characteristics of our dataset.
\subsection{Task Definition}
\label{sec: definition}
We define the task of I-VQA as: taking a video without explicit visual evidence $v~=~v_0 \setminus v_e$, a question $q$, and a corresponding answer set $\mathbb{A}$ as input, the model is required to deduce the correct answer ${a}^{\ast}$ from the answer set $\mathbb{A}$. 
Where $v_0$ indicates the complete video with a duration of $t$, and $v_e$ indicates the explicit visual evidence that occurs during  $\mathbb{T_e}=\{t_e^1, ..., t_e^n\}, where~t_e^i = [t_0, t_1], 0 \leq t_0<t_1 \leq t$. The reasoning process can be represented by: $ a^* = \arg \max _{a\in \mathbb {A}}~f_{w}(a|q, v, \mathbb {A})$, where $f_w$ represents a mapping function with learnable parameters $w$. 

Moreover, to ensure the implicitness, we appropriately extend the explicit visual evidence, where the certain excluded timestamp $t_e^i$ can be represented by: $t_e^i = [max(0, t_0-{T}/{\sigma}), min(t, t_1+{T}/{\sigma})]$, where $\sigma$ indicates a constant value. 
Besides, following previous VideoQA tasks~\cite{videollava,qwen,pllava,stllm}, we introduce the open-ended evaluation for a subset in the I-VQA task. 
After an extremely strict manual assessment, the open-ended evaluation subset is identified, which satisfies the condition of a unique correct answer without any imposed restrictions. The evaluation is represented by: $a_o = f_{w}(q, v)$. $a_o$ is the generated responses, which can be in free-form language for open-ended VideoQA. 

\begin{figure*}[!t]
\begin{center}
\includegraphics[width=\textwidth]{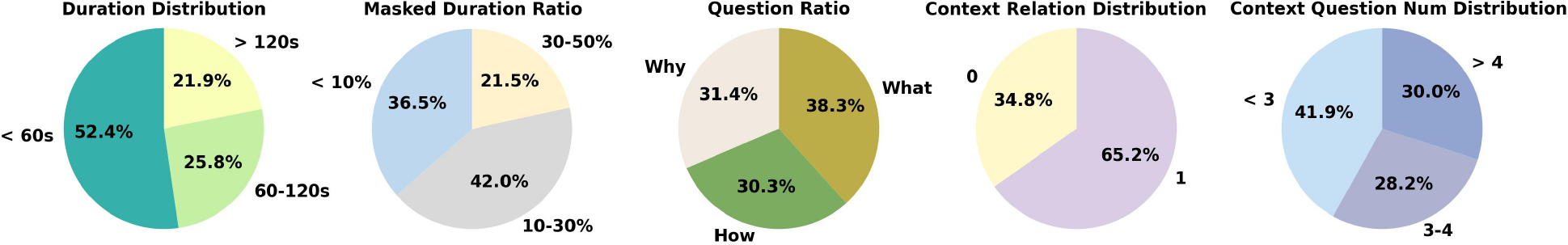}
\end{center}
\caption{\textbf{Dataset Statistics.} The majority of videos have a duration of around one minute, with explicit evidence typically accounting for 10-30\% of the video. Questions starting with `Why', `How', and `What' are common. Most of the context clues transformed from context QA pairs are beneficial for answering implicit questions, with the majority containing 3-4 context clues. }
\label{fig: statistics}
\end{figure*}

\subsection{Task Dataset}
\label{sec: dataset}
\noindent\textbf{Data Source.}
The I-VQA task involves question-answer pairs where key visual evidence is invisible. To set up the task setting for implicit questions, we systematically reorganize and leverage explicit visual evidence annotations from widely accepted Grounded VideoQA datasets, including Next-GQA~\cite{nextgqa}, E.T. Bench~\cite{etbench}, and REXTIME~\cite{rextime}. 
For the specific implementation, we slightly extend ($\sigma$ is set to 20) and manually mask these explicit visual evidence segments in videos.

Next-GQA comprises 8,911 QA pairs and 1,557 videos from NextQA~\cite{nextqa} which focuses on complex temporal and causal reasoning. 
E.T. Bench includes 10,000 grounded QA samples, requiring models to demonstrate event-level reasoning. 
REXTIME consists of 12,000 QA samples sourced from ActivityNet~\cite{acititynet} and QV-Highlights~\cite{qv-highlight} and is to evaluate the ability to perform temporal reasoning.\\

\noindent\textbf{Data Cleaning.} 
To ensure the necessity of explicit visual evidence and the uniqueness of the correct answer ${a}^{\ast}$ for the answer set $\mathbb{A}$, we conduct additional data filtering:

\textbf{1. We exclude questions that could be correctly answered purely through commonsense reasoning.} During the commonsense exclusion phase, only the questions, denoted as $q$, and their corresponding answer sets, $\mathbb{A}$, were provided for assessment. The primary objective was to eliminate instances that could be answered correctly without the associated visual input $v$. A quintessential example of such an excluded question is, ``Q: How does the boy feel after he has gotten up the first time he successfully rolled?~--~A: Happy.''
To do this, five annotators identified questions answerable by commonsense alone, and the Pearson Correlation of inter-annotator agreement achieves 0.95 (ranging from -1 to 1, more relative when closer to 1). 
Besides, our protocol was stringent: a question was removed if even one annotator flagged it. 
Finally, the effectiveness of this process was validated when GPT-4o, operating without visual input, performed at a level close to random chance on the test set, confirming the rigor of our screening~\cite{mecd+, egoschema}.

\textbf{2. We exclude questions focused on simple description questions and temporal location questions.} Descriptive questions refer to questions about non-inferential details, such as the color or quantity of something. 
The screening for these questions was conducted by GPT-4 on the input questions, and the rigor of this process was further ensured by manual review. 
Specifically, GPT-4 removed around 2,000 examples from 10,000. We verified that these removals met the criteria. Then, we conducted a further review of the remaining examples using cross-validation by five annotators. The Pearson Correlation between the annotators reached nearly 0.98. Consistent cases were excluded, and a meeting was held to decide on the removal of controversial examples, resulting in an additional 168 exclusions in total.
Temporal location questions, on the other hand, were filtered out through a string-matching method to remove QA pairs whose answers contained specific timestamps.

\textbf{3. We exclude questions that are insufficient to be reasoned out with contextual visual information.}
Insufficient answer exclusion refers to questions that cannot be deduced from context, such as those beginning with `When' asking for a time, or `Where' inquiring about a location. This exclusion was also implemented by a string-matching method.
Furthermore, we exclude questions that are manually identified as undeducible from context through a similar rigorous human examination; a detailed case is shown in the Appendix.Sec.~\ref{app_clean}.

These exclusions ensure that the correct answer $a^{\ast}$ of each question can be uniquely reasoned out, avoiding ambiguity in answers. 
Based on the three principles, we collect 5,549 QA samples from about 30,000 original QA samples, where 3,749 samples of I-VQA are for the training set and 1,800 samples are split into the test set.\\

\noindent\textbf{Data Annotation.} 
As discussed above, effectively leveraging clues is crucial when tackling implicit video questions. To support future research, we have annotated contextual clues for all implicit questions in the I-VQA training set.

The annotation process consists of two main steps. 
First, we generate numerous ``action-intent'' pairs. 
To do this, we rewrite the context question into a declarative sentence to serve as the ``action,'' and use the correct answer to the question as the ``intent,'' following the analysis in IntentQA~\cite{intentqa}. 
As indicated by the blue arrows in Fig.~\ref{fig: anno}, these contextual QA pairs have time durations that do not overlap with the target implicit question. 
This non-overlapping condition is imposed to \textbf{mitigate potential information leakage}, ensuring that the clues do not directly contain parts of the answer.

Next, we filter the generated ``action-intent'' pairs to select clues that are relevant to the implicit question. 
We annotate the relations $r = [ r_1, ..., r_n]$ between the context pair $1 \sim n$ and the implicit question $q$ using GPT-4, following the causal relation definition in REXTIME~\cite{rextime} and MECD~\cite{MECD}. 
Specifically, we determine whether a causal relation exists between the acquisition of context clues and the deduction of the implicit answer. Additionally, we instruct GPT-4 to pay attention to using the counterfactual inference while avoiding the illusory existence causality~\cite{MECD}.
We label context clues contributing to implicit question answering as ``1'' and those that don't as ``0''. Furthermore, we manually validate the quality of 200 causal relationship annotations and find that they highly align with human expert judgment~(major choice of 10 annotators with video understanding research experiments), with a similarity of over 92\% (which is robust enough for model learning to distinguish which clues contribute to implicit reasoning~\cite{MECD, rextime}).\\

\noindent\textbf{Data Statistics.} 
Data statistics are shown in sub-figures and the caption of Fig.~\ref{fig: statistics}. Most questions are centered around ``Why''~(\textit{e.g.}, ``Why did the man in black in the kitchen start laughing?''), ``How''~(\textit{e.g.}, ``How does the man engage with the women and showcase his culinary skills?''), and ``What''~(\textit{e.g.}, ``What did the ladies do after they passed the men some things?''). 
Our dataset highlights that each video contains multiple valuable context clues, which can serve as effective reasoning paths.
\section{Methods}
\label{sec:method}
\begin{figure*}[t]
\begin{center}
\includegraphics[width=0.98\textwidth]{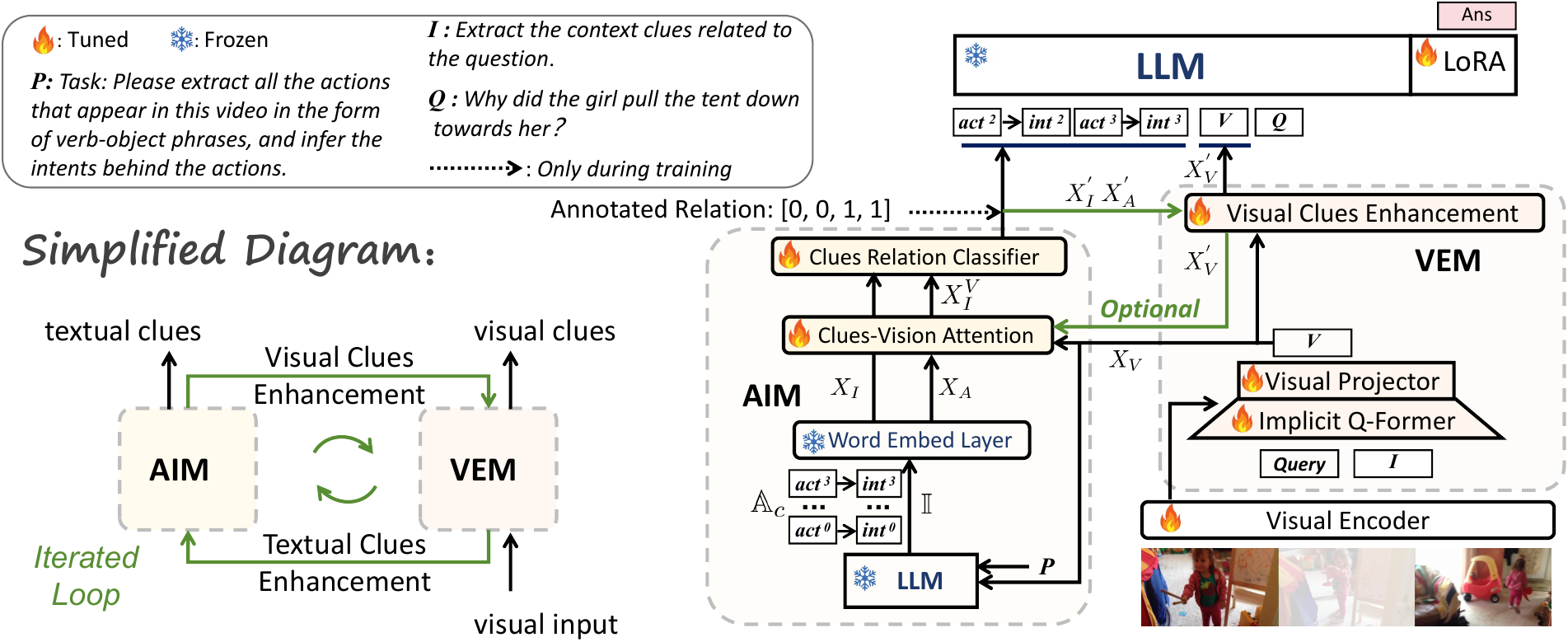}
\end{center}
\vspace{-5pt}
\caption{\textbf{IRM Framework.} IRM employs dual-stream modeling of contextual actions and intent clues as implicit reasoning chains. Dual clues are generated and refined by AIM through hallucination mitigation and clue-question relationship deduction. Meanwhile, VEM takes the refined clues to enhance visual input. Finally, LLM is prompted with multi-modal dual clues to conduct implicit reasoning.}
\label{fig: framework}
\end{figure*}

In this section, we present IRM as shown in Fig.~\ref{fig: framework}, which mainly consists of the visual encoder, AIM (Action-Intent Module)~(Sec.~\ref{sec: aim}), VEM (Visual Enhanced Module)~(Sec.~\ref{sec: vem}), and the LLM. The AIM module is responsible for inferring clues relevant to the implicit question, while VEM enhances the visual representations of the clues inferred by AIM. Finally, the multimodal clue representations output by both modules are fed into the LLM to derive the answer to the final question.

\subsection{Action Intent Module}
\label{sec: aim}


As discussed above, drawing on theories from cognitive science, we construct contextual clue information as an additional reasoning pathway to assist in implicit reasoning tasks. 
Furthermore, the visual evidence in the context is typically explicit. Consequently, extracting these direct, shallow intentions from actions is much less challenging than directly deducing the indirect, implicit intentions behind the question. 
Therefore, we proposed the AIM (Action Intent Module), which includes the extraction and refinement of Action Intent clues. 

For the extraction of clues, we prompt our model (which can be substituted with other MLLMs, such as GPT-4o, during training) to leverage the contextual visual information to extract this explicit evidence and formulate it into candidate contextual clues structured as "action-intent" pairs.
Specifically, we prompt it with instruction prompt \textit{P} (shown in the left-top legend of Fig.~\ref{fig: framework}:~\textit{P: Task: Please extract ...}) to generate action-intent clue candidates. 

LLM is prompted with the uniformly sampled frames of implicit visual input $v$ and the In-Context Learning prompt. Context action sets $\mathbb{A}_c =\{ act^0, ..., act^n\}$ and corresponding intents set $\mathbb{I} =\{ int^0, ..., int^n\}$ are generated, which are then embedded and concatenated as follows:
\begin{equation}
\begin{split}
X_A &= \texttt{Cat}(f_T(act^0), ..., f_T(act^n)), \\
X_I &= \texttt{Cat}(f_T(int^0), ..., f_T(int^n))
\end{split}
\end{equation}
where $f_T$ indicates the word embedding layer, $\texttt{Cat}$ indicates concatenate. 

\textbf{To further refine the clues, we focus on mitigating the hallucination problem of clues while reasoning out key clues beneficial for implicit reasoning.}
The current action clues may be influenced by hallucinations, but this can be effectively mitigated through careful verification using visual information~\cite{MECD, hallucination, mmvu}.
Therefore, we incorporate cross-action-vision attention to query each action candidate with global visual information $Q_V$ extracted by QFormer. 
Specifically, the mechanism is performed by the following equation:
\begin{equation}
X_I^V = \text{$\Phi_{h}^C$}(X_A, f_P(Q_V))
\label{eq5}
\end{equation}
where $\Phi_{h}^C$ indicates the cross-attention for hallucination mitigating, the former input of  $\Phi_{h}^C$ indicates Query, while the latter is for Key and Value. $f_P$ indicates the visual projection layers. By further verifying clues with global visual information, we enhance the injection of temporal awareness while sequentially identifying contextual clues to mitigate hallucinations.

Moreover, action-intent clue pairs contribute differently to the reasoning of the implicit question. 
Therefore, we introduce the causal relations classifier with implicit question $q$ to determine which intent-action pairs assist in reasoning.
Specifically, we concatenate the visually enhanced $X_I^V$ and the original intent tensor $X_I$ with the embedded question $f_T(q)$ and then pass them to the causal attention classifier~\cite{MECD} for relationship deduction. 
This process is represented as:
\begin{equation}
\begin{aligned}
X_I^Q &= \texttt{Cat}(X_I, f_T(q)), X_I^{V,Q} = \texttt{Cat}(X_I^V, f_T(q)) \\
\hat{r} &= g_{r}(\texttt{{Cat}}(\text{$\Phi_{r}^C$}(X_I^Q, X_I^{V,Q}), \text{$\Phi_{r}^S$}(X_I^Q)))
\end{aligned}
\end{equation}
where $\Phi_{r}^C$ and $\text{$\Phi_{r}^S$}$ represents cross-attention and self-attention, and $g_{r}$ represents the linear layer. Relational reasoning is achieved through causal attention mechanisms, which model the deep causal relationship between each clue and the question—that is, whether the clue contributes to answering the question.
We introduce the relation loss $\mathcal{L}_R$ to supervise the reasoning process during training:
\begin{equation}
    \mathcal{L}_R = \sum_{j=1}^{M} \mathcal{L}_{ce}(\hat{r}_{\hat{\sigma}(j)}, r_j)
\end{equation}
where $\mathcal{L}_{ce}$ is the cross-entropy loss. Since the set of $N$ predictions, $\{\hat{r}_i\}_{i=1}^N$, and the set of $M$ ground truth relation annotations, $\{r_j\}_{j=1}^M$, do not have a one-to-one correspondence, we first find an optimal bipartite matching between them. This matching, denoted as $\hat{\sigma}$, is found using an algorithm like the Hungarian algorithm to minimize the overall cost of pairing predictions to ground truths. The final loss is then the sum of the cross-entropy losses computed only over these optimally matched pairs, where $\hat{r}_{\hat{\sigma}(j)}$ is the prediction assigned to the ground truth $r_j$. 
During inference, there is no need to match against annotations; instead, the clue candidates are directly refined. 
The refined context action clues $X_A'$ and intent clues $X_I'$ can be represented as follows:
\begin{equation}
\begin{split}
X_A' = \texttt{Cat} \left( {X_A}[i] \right), X_I' = \texttt{Cat} \left( {X_I}[i] \right), \\ 
\hat{r}[i, 0] > \hat{r}[i, 1].
\end{split}
\end{equation}
By retaining only the clues inferred to be relevant to the question, AIM effectively filters out Action-Intent clues that are semantically accurate and conducive to implicit reasoning.

\subsection{Visual Enhanced Module}
\label{sec: vem}
As discussed above, the VEM first introduces QFormer~\cite{blip2} to compress extensive visual input $f_V(v)$ from the visual encoder, represented as $Q_V$. 
During the compression, the instruction to Qformer is: \textit{``Extract the context clues related to the question: q''}.
Following the previous works~\cite{videochat2, videollava, videollama, compress}, compressed visual input $Q_V$ is then sent to visual projection layers $f_P$ to align with textual embeddings, represented by $X_V=f_P(Q_V)$.

The aligned visual input is then sent to the AIM module for dual context clues reasoning. 
Meanwhile, the current visual extraction $X_V$ only implicitly suggests the need to extract key context visual clues.
Therefore, for explicit visual enhancement, we establish a feedback path from AIM to VEM, introducing refined context clues generated by AIM to enhance the vanilla visual representation $X_V$ through a cross-attention mechanism:
\begin{equation}
X_V' = \text{$\Phi_{v}^C$}(X_V, \texttt{Cat}(X_A', X_I')).
\label{eq9}
\end{equation}
where $\Phi_{v}^C$ indicates the cross-attention. Therefore, by further querying the selected key textual clues within the visual input, clue-enhanced visual embeddings are obtained. 
Additionally, as shown in the simplified diagram in Fig.~\ref{fig: framework}, the coupled design of AIM and VEM enables additional enhancement iterations during inference. 
Specifically, the global visual input $f_P(Q_V)$ of Eq.~\ref{eq5} can be replaced by enhanced $X_V'$ after Eq.~\ref{eq9} through an additional iteration. 
The LLM finally conducts reasoning:
\begin{equation}
Ans = \text{LLM}(X_A', X_I', X_V', q).
\end{equation}
In this way, the model effectively utilizes both textual and visual clues of context when answering implicit questions. Detailed prompts can be referred to in the Appendix.Sec.~\ref{app_prompt}.
\section{Experiments}
\label{sec: exp}

\subsection{Implementation Details}
\label{sec: details}
\noindent\textbf{Architectures and Training.} We utilize VideoChat2~\cite{videochat2} as baseline with UMT-L~\cite{umt} as visual encoder and Mistral-7B~\cite{mistral} as LLM. 
We initialize IRM with the pre-trained VideoChat2 model after vision-language alignment, followed by an SFT stage to enhance the model's implicit reasoning abilities.
We trained IRM on 8 A6000 GPUs for about 3 hours. 
\textbf{We use GPT-4o by default for candidate clue generation during the training phase of IRM. During inference, however, we use the same clue extraction prompt to have the IRM model generate the clues itself.}
More details are in the Appendix.Sec.~\ref{app_detail}.\\
\noindent\textbf{Datasets.}
We conduct experiments on the proposed I-VQA dataset and a representative advertising dataset, the PSAV dataset~\cite{adsdataset}. 
Advertising datasets are particularly well-suited for implicit reasoning, as they often employ subtle strategies to persuade the audience rather than explicitly instructing them to purchase a product or accept a premise. 
The PSAV dataset offers a curated collection of 1,002 video advertisements for analyzing persuasive tactics. Each video is annotated to identify the effective ones of 12 distinct advertising persuasion strategies.

\noindent\textbf{Evaluation Metrics.} For the evaluation of implicit video question answering, we follow the mainstream paradigms for video QA: open-ended and multiple-choice evaluation. For the open-ended setting of I-VQA, we adopt the ``Score'' and ``$\text{Acc}_\text{open}$'' metric following previous VideoQA works based on LLMs~\cite{videollava,qwen,pllava,stllm}, where prediction is scored~(from 0 to 5) and judged true or false by GPT-3.5. Besides, the high coherence between ``Score'' and human is proved in the Appendix.Sec.~\ref{app_score}, with Pearson correlation coefficient~\cite{pearson} reaching a high score of 0.93. 
We adopt option ``Accuracy'' for multi-choice setting. 
For PSAV, we follow the metric in~\cite{adsdataset}: ``Accuracy'' to evaluate the answer precision. 
Moreover, since multiple correct answers may exist for PSAV, we introduce the complement metric ``Recall'' to evaluate the proportion of the answers covered.

\begin{figure*}[t]
\centering
\begin{minipage}[t]{0.51\textwidth}
    \centering
     \captionof{table}{\textbf{Results on I-VQA.} IRM reaches SOTA performances in both multi-choice and open-ended settings. Clues are fairly provided by default; $*$ indicates reasoning without clues provided. ZS indicates zero-shot, FT indicates fine-tuning.}
    \label{tab: mainresults}
    \resizebox{1.00\textwidth}{!}{
      \setlength{\tabcolsep}{0.4mm}{
      \begin{tabular}{ccccc}
        \toprule
        \multirow{2}{*}{\textbf{Paradigm}} &\multirow{2}{*}{\textbf{Method}} & \textbf{Multi-choice} &\multicolumn{2}{c}{\textbf{Open-ended}} \\
        & & \textbf{Accuracy} &\textbf{Score} &\textbf{$\text{Acc}_\text{open}$} \\
        \midrule
       - & Random Guess &21.14 &- &-\\
        \midrule
        \multirow{15}{*}{$\text{\textbf{ZS}}$} 
        &\textit{Reasoning LLM}\\
        &DeepSeek-R1~\cite{deepseek} &49.75  &3.23  &54.09 \\
        &OpenAI-o3~\cite{o3} &53.77 &3.20   &\cellcolor{green!8}\textbf{55.35}  \\
        &\textit{Proprietary ImageLLM}\\
        & $\text{GPT-4o}^*$~\cite{gpt4} &\cellcolor{red!5}50.40 &\cellcolor{red!5}3.25 &\cellcolor{red!5}52.66\\
        & Gemini-2.5-Pro~\cite{gemini} &52.84 &3.25  &53.57 \\
        & Claude-3.7-Sonnet~\cite{claude} &53.31 &3.27  &52.83 \\
        & GPT-4o~\cite{gpt4} &\cellcolor{green!3}54.38 &\cellcolor{green!8}\textbf{3.28} &54.41 \\
        \cmidrule{2-5}
        & \textit{Open-Source MLLM}\\
        &Minigpt4-video~\cite{Minigpt4-video} &35.85 &2.88 &41.60\\
        & VideoLLaVA~\cite{videollava}  &36.26 &2.94 &42.80\\
        & PLLaVA~\cite{pllava} &44.38 &3.11 &47.93\\
        &Qwen2.5-VL-7B~\cite{qwen} &46.95 &3.13   &48.72 \\
        &NViLA~\cite{nvila} &48.42 &2.95 &46.08 \\
        & VideoChat2~\cite{videochat2} &\cellcolor{yellow!20}50.27 &\cellcolor{yellow!20}3.15 &\cellcolor{yellow!20}52.24\\
        \midrule
        \multirow{7}{*}{$\text{\textbf{FT}}$} 
        & \textit{Open-Source MLLM}\\
        &PLLaVA~\cite{pllava} &43.82 &3.17  &52.46 \\
        &Qwen2.5-VL-7B~\cite{qwen} &51.26 &3.26 &54.32 \\
        & VideoChat2~\cite{videochat2} &51.90 &3.27 &53.79\\
        \cmidrule{2-5}
        & \textbf{Self-IRM} &\cellcolor{green!8}\textbf{54.50} &\cellcolor{green!8}\textbf{3.28} &\cellcolor{green!3}55.31 \\
        &  \textbf{{IRM}} &\cellcolor{green!15}\textbf{{\underline{55.14}~(+4.9)}} &\cellcolor{green!15}\textbf{{\underline{3.29}}} &\cellcolor{green!15}\textbf{\underline{56.02}}\\
        \midrule
        - & Human &92.50  &- &-\\
        \bottomrule
      \end{tabular}
    }}
\end{minipage}
\hfill
\begin{minipage}[t]{0.47\textwidth}
    \centering
    \vspace*{0.8\baselineskip}
    
    \includegraphics[width=0.8\textwidth]{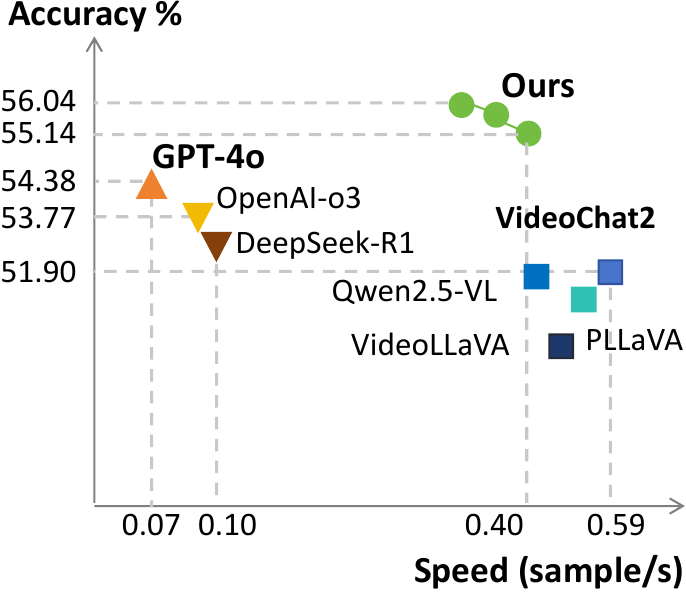}
    \caption{\textbf{Accuracy and Efficiency Trade-off Results.} IRM achieves both optimal performance and high efficiency during reasoning.}
    \label{fig: tradeoff}
    
    \includegraphics[width=0.99\textwidth]{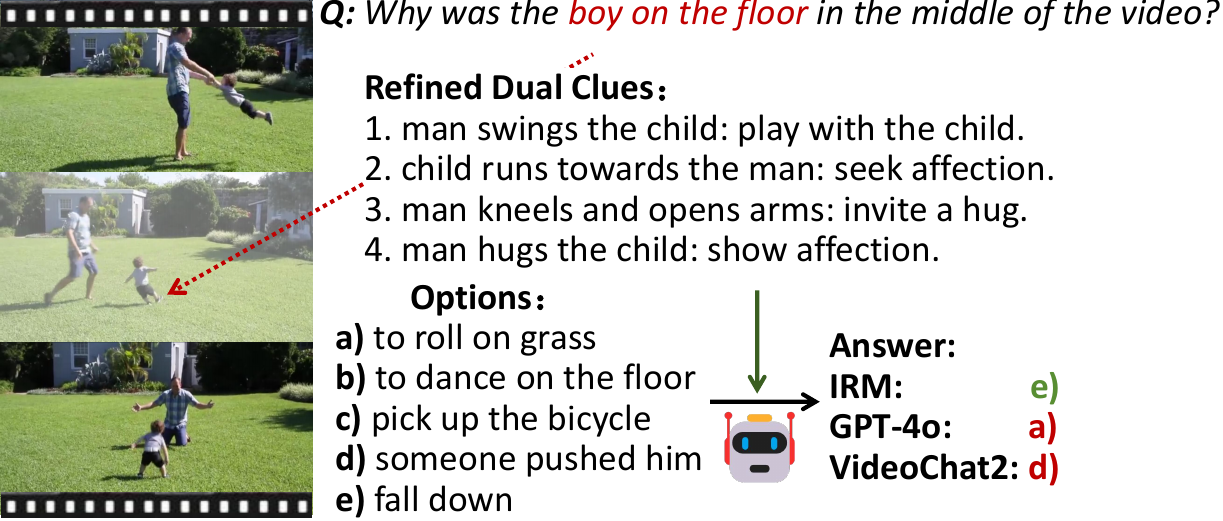}
    \captionof{figure}{\textbf{Visualizations of multi-choice setting of I-VQA.} The result indicates the powerful reasoning and clues refining ability of IRM.}
    \label{fig: vis-mc}
\end{minipage}
\end{figure*}

\noindent\textbf{Comparative Methods.}
We compare IRM with powerful Proprietary ImageLLMs GPT-4o~\cite{gpt4o}, Gemini-2.5-Pro~\cite{gemini}, and Claude-3.7-Sonnet~\cite{claude}.
Similarly, we evaluate the leading open-source MLLMs, VideoLLaVA~\cite{videollava}, PLLaVA~\cite{pllava}, Minigpt4-video~\cite{Minigpt4-video}, NViLA~\cite{nvila}, Qwen2.5-VL~\cite{qwen}, and VideoChat2~\cite{videochat2}. We also evaluate reasoning models OpenAI-o3~\cite{o3} and DeepSeek-R1~\cite{deepseek}.
For fairer comparison, we fine-tune the represented open-source MLLMs PLLaVA, Qwen2.5-VL, and VideoChat2 when conducting comparisons with our proposed IRM. 
\textbf{To ensure fairness, all comparison models are also prompted with the same context clues generated by GPT-4o during reasoning, which is strictly aligned with our IRM}. Fine-tuning details and prompts for IRM and other models are in the Appendix.Sec.~\ref{app_prompt}, specific API version details are in the Appendix.Sec.~\ref{app_detail}.

\subsection{Main Results}
\label{sec: main results}
\noindent\textbf{Evaluation of the I-VQA Dataset.}\\
As shown in Tab.~\ref{tab: mainresults}, our IRM achieves SOTA in both the multi-choice and open-ended settings of the I-VQA dataset, with an accuracy of 55.14 and a GPT score of 3.29. 
This surpasses existing MLLMs, including fine-tuned VideoChat2~\cite{videochat2}, as well as GPT-4o~\cite{gpt4}. 
Specifically, IRM demonstrates improvements of over 0.67\% in $\text{Acc}_\text{open}$ and 0.76\% in accuracy compared to the closest OpenAI-o3 and GPT-4o, exhibiting slightly stronger implicit reasoning abilities than strong commercial models. 
Accuracy and efficiency trade-off is shown in Fig.~\ref{fig: tradeoff}, and detailed statistics are in the Appendix.Sec.~\ref{app_speed}. The results indicate that IRM achieves both optimal performance and high efficiency.
Additionally, IRM with self-generated clues during both training and evaluation achieves the second-best performance, indicating that even under resource constraints and without relying on commercial MLLMs, the proposed clue-based strategies can effectively enhance the model's implicit reasoning ability.

\begin{figure*}[t]
\centering
\begin{minipage}[t]{0.50\textwidth}
    \vspace*{0.8\baselineskip}
    
    \includegraphics[width=\textwidth]{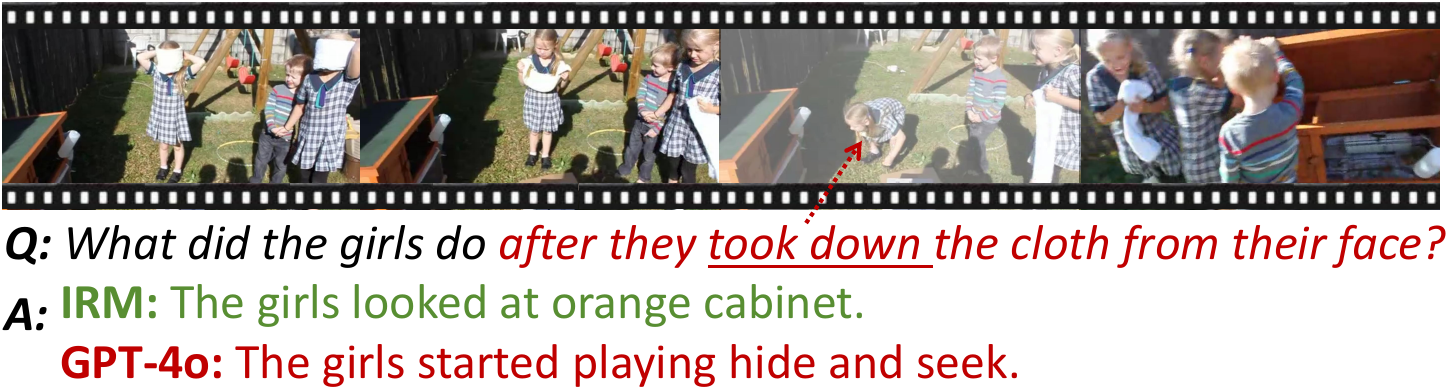}
    \caption{\textbf{Visualization of the Open-ended Setting of I-VQA.} GPT-4o linked removing the blindfold to starting hide-and-seek, missing the key clue that no one had disappeared, while IRM correctly reasoned the status.}
    \label{fig: vis-open}
    
    \vspace{5pt} 
    
    \includegraphics[width=\textwidth]{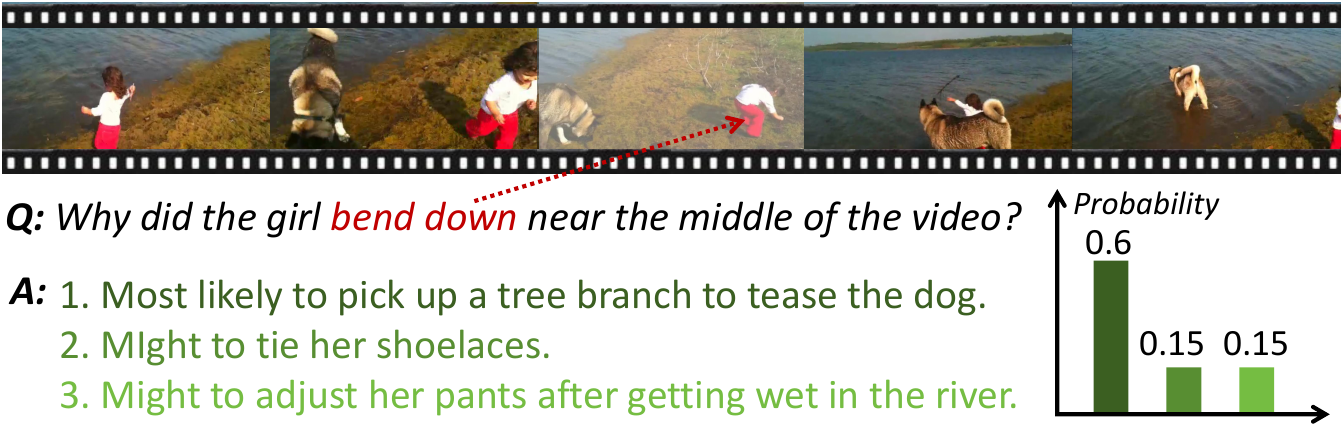}
    \caption{\textbf{Open-ended Result of Example with Non-unique Answers.} In cases where hidden information has multiple possibilities, IRM effectively generates multiple answers with associated probabilities.}
    \label{fig: vis-multi}
\end{minipage}
\hfill
\begin{minipage}[t]{0.47\textwidth}
    \centering
    \captionof{table}{\textbf{Results on PSAV Dataset.} IRM achieves SOTA performances in zero-shot test on unseen advertisement tasks.}
    \label{tab: ads}
    \resizebox{\textwidth}{!}{
      \setlength{\tabcolsep}{0.3mm}{
      \begin{tabular}{ccc}
        \toprule
        Method & Recall & Accuracy \\
        \midrule
        Story + Flan-t5 Classifier~\cite{adsdataset} &- &33.41 \\
        Story + GPT-3.5 Classifier~\cite{adsdataset} &- &35.02 \\
        OpenAI-o3~\cite{o3} &36.38 &\cellcolor{green!5}45.84 \\
        GPT-4o~\cite{gpt4} &33.51 &44.92 \\
        \midrule
        DeepSeek-R1~\cite{deepseek} &23.62 &21.96 \\
        VideoChat2~\cite{videochat2} &\cellcolor{yellow!20}20.92 &\cellcolor{yellow!20}39.22 \\
        PLLaVA~\cite{pllava} &\cellcolor{green!3}38.36 &39.40 \\
        NViLA~\cite{nvila} &38.13 &41.09 \\
        \midrule
        \textbf{Self-IRM} &\cellcolor{green!5}\textbf{39.02} &\cellcolor{green!3}\textbf{45.14} \\
        \textbf{IRM} &\cellcolor{green!15}\textbf{45.41~(+24.5)} &\cellcolor{green!15}\textbf{46.08~(+6.9)} \\
        \bottomrule
      \end{tabular}
    }}

    \vspace{10pt}

    \includegraphics[width=\textwidth]{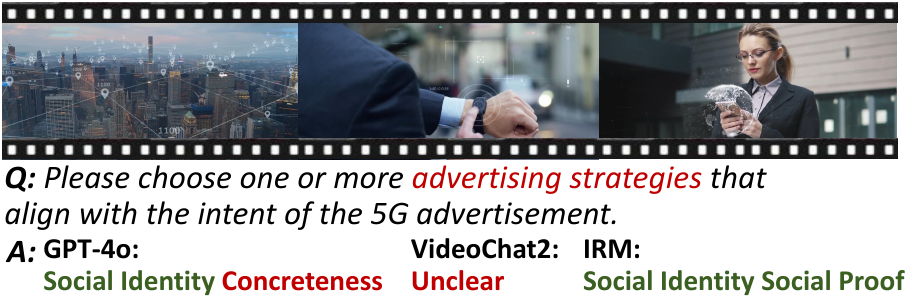}
    \captionof{figure}{\textbf{Visualizations of the PSAV Dataset.} The results show that IRM accurately grasps the persuasion strategies.}
    \label{fig: vis-ivqa}
\end{minipage}
\end{figure*}



The performances of $\text{GPT-4o}^*$~(GPT-4o without context clues provided, marked in red in Tab.~\ref{tab: mainresults}) indicate the challenge of the implicit reasoning task.
Besides, the fine-tuned MLLMs did not show a significant improvement over the zero-shot models, indicating that guiding the model's reasoning process is far more crucial than simply exposing it to implicit questions. 
Conversely, our IRM significantly improves accuracy by 4.87\% compared to the baseline, indicating the effectiveness of enhanced dual context clues. 

Reasoning LLM OpenAI-o3 and DeepSeek-R1 demonstrate strong reasoning abilities under the enhancement of contextual clues generated by GPT-4o, successful and failed examples are in the Appendix.Sec.~\ref{app_r1}. 
For validating the exclusion of commonsense questions, we also evaluate two reasoning models only with the question and options, the accuracy drops to 23.51 and 22.74, which is only slightly higher than random guessing. This result also indicates a minimal language bias~\cite{perception} of I-VQA.
Visualizations of different settings in Fig.~\ref{fig: vis-mc} and Fig.~\ref{fig: vis-open} demonstrate that IRM successfully infers the correct answer with dual refined clues. 
We also present the subjective results under the open-ended condition for examples that are not included in the open-ended subset. 
As shown in Fig.~\ref{fig: vis-multi}, IRM effectively generates multiple reasonable outcomes along with their probabilities.\\


\noindent\textbf{Evaluation of the PSAV Dataset.}\\
As detailed in Tab.~\ref{tab: ads}, the IRM model has established a new state-of-the-art (SOTA) in the zero-shot generalization evaluation on the PSAV dataset. 
These results represent a significant improvement over VideoChat2, with IRM outperforming it by 24.49\% in recall and 6.86\% in accuracy.
Although there are no video examples related to the advertising domain in the training set, we still achieved SOTA performance, demonstrating that our proposed method effectively enhances the reasoning capacity for general implicit questions. Visualizations are shown in Fig.~\ref{fig: vis-ivqa} and the Appendix.Sec.~\ref{app_vis}.

\subsection{Ablation Study}

\label{sec: ablation}
\begin{figure*}[t]
\centering
\begin{minipage}[t]{0.48\textwidth}
    \centering
    \captionof{table}{\textbf{Ablation Study.} We validate the effectiveness of the basic dual clues structure and confirm the refinement and enhancement effect of the proposed coupling AIM and VEM modules.}
    \label{tab: ablation}
    \resizebox{\textwidth}{!}{
    \setlength{\tabcolsep}{1.0mm}
    \begin{tabular}{ccc|ccccc}
        \toprule
         \multicolumn{3}{c}{$\textbf{Context Clues Designs}$} & \multicolumn{3}{c}{$\textbf{IRM Designs}$} 
         &\multirow{2}{*}{Acc} \\
         Clue & Tracking &Scene Graph & Instruction & AIM & VEM  \\
        \midrule
         &  & & & & &51.94 \\
         &\checkmark  & & & & &52.61 \\
         & & \checkmark & & & &52.78 \\
        \cellcolor{green!5}\checkmark & & & & & &\cellcolor{green!5}\textbf{53.28} \\
        \midrule 
        \checkmark & & &\checkmark & & &53.50 \\
        \checkmark & & &\checkmark & \checkmark & &53.67 \\
        \checkmark & & &\checkmark &  &\checkmark &53.00 \\
        \checkmark & & & \checkmark & \cellcolor{green!5} \checkmark &\cellcolor{green!5}\checkmark &\cellcolor{green!5}\textbf{55.14} \\
        \bottomrule
    \end{tabular}
    }
    
    \vspace{10pt}

    \centering
    \captionof{table}{\textbf{Visual Encoder and LLM Ablations.} The greater impact of encoder indicates the importance of visual clues.}
    \resizebox{\textwidth}{!}{
      \setlength{\tabcolsep}{3.3mm}{
      \begin{tabular}{ccc}
        \toprule
        Visual Encoder &LLM &Acc \\
        \midrule
        EVA-CLIP-g~\cite{eva} &Vicuna-7B v1.5~\cite{vicuna} &51.69 \\
        EVA-CLIP-g~\cite{eva} &Mistral-7B ~\cite{mistral} &51.98\\
        UMT-L~\cite{umt} &Vicuna-7B v1.5~\cite{vicuna} &54.31 \\
        UMT-L~\cite{umt} &Mistral-7B~\cite{mistral} &55.14 \\
        \bottomrule
      \end{tabular}
    }}
    \label{tab: encoderllm}
\end{minipage}
\hfill
\begin{minipage}[t]{0.48\textwidth}
    \centering
    \vspace*{1.0\baselineskip}
    
    \begin{tikzpicture}[scale=0.7]
    \begin{axis}[
        xlabel={Percent~(Unit 1 for 3.7k Data / 32 Frames / 5 Iters)},
        ylabel={Accuracy (\%)},
        xlabel style={yshift=2pt}, 
        ylabel style={yshift=-3pt}, 
        xtick={0.2,0.3,0.4,0.5,0.6,0.7,0.8,0.9,1.0},
        ytick={48,51,54,57},
        ymin = 48,
        ymax = 57,
        xmin = 0.18,
        xmax = 1.02,
        legend style={at={(0.98,0.98)}, anchor=north east, legend columns=2},
        grid=both,
        grid style={line width=0.5pt, draw=gray!50},
        major grid style={line width=0.3pt,draw=gray!90},
        width=9.0cm,
        height=4.0cm
    ]

    \addplot[color=blue,mark=*] coordinates {
        (0.25, 55.14)
        (0.5, 54.95)
        (1.0, 56.15)
    };

    \addplot[color=red,mark=square*] coordinates {
        (0.2, 48.85)
        (0.4, 49.45)
        (0.6, 52.60)
        (0.8, 54.50)
        (1.0, 55.14)
    };

    \addplot[color=green,mark=triangle*,mark size=3] coordinates {
        (0.2, 55.14)
        (0.4, 55.59)
        (0.6, 56.04)
        (0.8, 56.28)
        (1.0, 56.35)
    };

    \addplot[domain=0.18:1.02, color=black, dashed] {53.50};
    \end{axis}
    \end{tikzpicture}
    \captionof{figure}{\textbf{Training Data, Reasoning Time and Input Frames Ablation Study.} Inference accuracy increases with more training data~\textcolor{red}{(red)} and more reasoning time~\textcolor{green}{(green)}. Optimal performance is reached with 8 input frames~\textcolor{blue}{(blue)} considering efficiency.}
    \label{quant}
    
    \vspace{5pt}

    \centering
    \includegraphics[width=\textwidth]{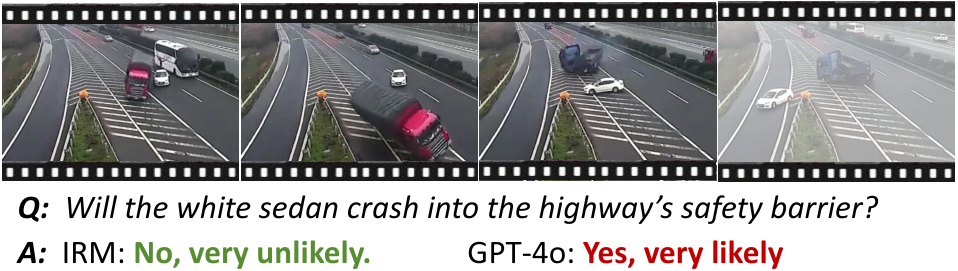}
    \captionof{figure}{\textbf{Visualization of Traffic-VQA.} The correct answer, ``most likely not'' is inferred by our IRM with context clues: the trucks braking to avoid the white.}
    \label{fig: vis-traffic}
\end{minipage}
\end{figure*}

\noindent\textbf{Experiments of the Dual Clues and Architecture.}
To demonstrate the effectiveness of incorporating context clues, we compare the performances between IRM and the following methods:
(1) No guidance.
(2) Utilizing Grounding-Dino~\cite{groundingdino} for object tracking.
(3) Employing PySceneDetect~\cite{PySceneDetect} for scene detection, followed by SceneGraphParser~\cite{SceneGraphParser} to convert the context scenes into a triplet-based graph. 
As shown in Tab.~\ref{tab: ablation}, context dual clues lead to improvements of 3.20\%, 2.53\%, and 2.36\%, respectively. 
This also proves that the SOTA performance of IRM is not simply achieved through video enhancement methods like partial masking~\cite{videomae}. 

Besides, we evaluate our design of VEM for enhancing visual information extraction. 
We also evaluate the coupled AIM and VEM, while the improvement is modest when introduced individually, combining both modules results in a significant boost. \\

\noindent\textbf{Experiments of Data, Reasoning Time and Frames.}
As shown in Fig.~\ref{quant}, accuracy continues to rise while increasing the data volume during SFT, similar to the post-training scaling law~\cite{aligner, scaling}. 






We also study the relationship between reasoning time and accuracy during inference. By iteratively enhancing the visual and textual clues, as the green line in Fig.~\ref{quant}, performance improves with the additional 1-3 iterations~(from 55.14 to 55.59, 55.73, 56.04, indicating the inference scaling law with reasoning time~\cite{time, scaling}. 
Specifically, increasing iterations results in a better performance compared to traditional reasoning methods such as Greedy Search~\cite{greedy} of 55.65 and Majority Voting~\cite{vote} of 55.87 while maintaining the same FLOPs. 
Moreover, IRM reaches optimal performance with 8 frames considering efficiency, which may be related to the pre-training process.\\


\begin{figure*}[t]
\centering

\begin{minipage}[t]{0.48\textwidth}
    \centering
    \captionof{table}{\textbf{Robustness to Noisy Clues.} The minimal performance drop of IRM during noisy situations highlights the effectiveness of the clue refinement module.}
    \label{tab:robustness_test}
    \resizebox{\linewidth}{!}{%
      \setlength{\tabcolsep}{0.5mm}{
      \begin{tabular}{lccc}
        \toprule
        {Model} & Vanilla Acc. & Acc. with Noisy Clues & Acc. Drop \\
        \midrule
        VideoChat2 & 51.90 & 46.54 & 5.36 \\
        \textbf{IRM} & \textbf{55.14} & \textbf{53.83} & \cellcolor{green!10}\textbf{1.31} \\
        \bottomrule
      \end{tabular}%
      }}

 \vspace{15pt} 
    
    \captionof{table}{\textbf{Performance Breakdown by Question Type.} IRM's advantage is most pronounced on ``How'' questions, which require more complex reasoning.}
    \label{tab:qtype_breakdown}
    \centering
    \resizebox{\linewidth}{!}{%
      \setlength{\tabcolsep}{2mm}{
      \begin{tabular}{lccc}
        \toprule
        {Question Type} & {IRM Acc} & {GPT-4o Acc} & {Difference} \\
        \midrule
        Why     & 57.39 & 57.25 & \textbf{-0.14} \\
        What    & 54.82 & 54.69 & \textbf{-0.13} \\
        How     & 51.85 & 50.17 & \cellcolor{green!10}\textbf{+1.68} \\
        \bottomrule
      \end{tabular}%
    }}
\end{minipage}
\hfill
\begin{minipage}[t]{0.48\textwidth}
    \centering
    \captionof{table}{\textbf{Downstream Task Results of the Traffic-VQA Dataset.} In traffic-VQA with a certain degree of implicit meaning, IRM also shows strong generalization ability.}
    \label{tab: traffic}
    \resizebox{\textwidth}{!}{
      \setlength{\tabcolsep}{8.0mm}{
      \begin{tabular}{cc}
        \toprule
        Method &Accuracy\\
        \midrule
        \textit{Proprietary LLM}\\
        Claude-3.5-Sonnet~\cite{claude} &40.50\\
        Gemini-1.5-Pro~\cite{gemini} &44.96\\
        GPT-4o~\cite{gpt4} &45.83 \\
        OpenAI-o3~\cite{o3} &\cellcolor{green!8}47.15 \\
    
        \midrule
        \textit{Open-Source LLM}\\
        Minigpt4-video~\cite{Minigpt4-video} &30.71 \\
        VideoLLaVA~\cite{videollava}  &32.41 \\
        PLLaVA~\cite{pllava} &37.45 \\
        VideoChat2~\cite{videochat2} &\cellcolor{yellow!20}43.76 \\
        DeepSeek-R1~\cite{deepseek} &45.81 \\
        NViLA~\cite{nvila} &46.12 \\

        \midrule

       \textbf{Self-IRM} &\cellcolor{green!3}\textbf{46.22}\\
      \textbf{{IRM}} &\cellcolor{green!15}\textbf{{47.94}~(+4.18)} \\
       
        \bottomrule
      \end{tabular}
    }}
\end{minipage}
\end{figure*}

\noindent\textbf{Experiments of Visual Encoder and LLM.}
We default to utilizing UMT-L and Mistral-7B as our visual encoder and LLM, respectively, while also evaluating other models such as EVA-CLIP~\cite{eva} for the encoder and Vicuna~\cite{vicuna} for the LLM.
As shown in Tab.~\ref{tab: encoderllm}, the clues extraction ability of the visual encoder has a greater impact. 
This shows that, in implicit reasoning, the context clues, which serve as additional textual information and enable the extraction of visual clues, are more important than the subsequent reasoning.\\

\noindent\textbf{Experiments of Noisy Situation Evaluations.}
To evaluate the robustness of our framework against irrelevant or distracting information, we conducted an experiment simulating scenarios with noisy contextual clues. Specifically, we augmented the set of potential clues by 50\%, introducing noisy candidates randomly sourced from unrelated videos. This setup is designed to test the efficacy of our clue selection mechanism (the AIM module) in complex environments.

We compare the performance of our full IRM against the VideoChat2 baseline, which lacks a specialized clue-filtering mechanism. The results are presented in Tab.~\ref{tab:robustness_test}. The data shows that while the baseline model's accuracy drops substantially by 5.36\% in the presence of noisy clues, our IRM's performance degrades by only 1.31\%. 

This quantitative experiment validates the effectiveness of our proposed AIM module in identifying and filtering out irrelevant information. The minimal performance drop demonstrates the robustness of the IRM framework in handling complex and noisy video contexts, a crucial capability for real-world applications. These results strongly support our claim that the proposed approach is robust, especially when navigating complex situations with many potential distractors.


\subsection{Analysis of Performances of Different Types of Questions}

To further investigate the implicit reasoning capabilities of our model, we conduct a fine-grained performance analysis by breaking down the results based on the type of question. We evaluate IRM and the strong baseline, GPT-4o, on three primary implicit question categories: ``Why'', ``What'', and ``How''. The detailed accuracies are presented in Tab.~\ref{tab:qtype_breakdown}.

The results indicate that while both models perform comparably on ``Why'' and ``What'' questions, with only marginal differences, IRM exhibits an obviously larger advantage on ``How'' questions. Specifically, IRM outperforms GPT-4o by 1.68\% in this category, an advantage more than ten times greater than on the other two types. This finding is particularly salient, as ``How'' questions often necessitate more complex, multi-step causal reasoning. 
The substantial performance gap in this category provides strong quantitative evidence that IRM's architecture offers a distinct advantage in handling intricate reasoning tasks. 

\subsection{Generalization Evaluation}
 \begin{table*}[t]
  \centering
      \caption{\textbf{MVbench Subset Evaluation Results.} IRM model brings a general improvement in tasks such as action evolution and behavior prediction in video understanding. Overall, the average accuracy increases by 1.88, demonstrating strong robustness in general video reasoning.}
  \label{tab: mvbench}
  \begin{tabular*}{\dimexpr 0.88\textwidth}{@{\extracolsep{\fill}}clllll} 
    \toprule
    \textbf{Model} & \textbf{AP} & \textbf{MD} & \textbf{ER} & \textbf{OE} & \textbf{Ave} \\
    \midrule
    \textbf{VideoChat2}~\cite{videochat2} & 47.5 & 23.0 & 40.5 & 58.0 & \textbf{42.25} \\
    
    \multirow{2}{*}{\textbf{IRM}} & \cellcolor{green!8}56.1 $\uparrow$ & \cellcolor{green!8}27.1 $\uparrow$ & \cellcolor{red!5}37.8 & \cellcolor{red!5}55.5 & \cellcolor{green!8}\textbf{44.13}$\uparrow$ \\
    & \cellcolor{green!8}+8.6 & \cellcolor{green!8}+4.1 & \cellcolor{red!5}-2.8 & \cellcolor{red!5}-2.5 & \cellcolor{green!8}\textbf{+1.88} \\
    \bottomrule
  \end{tabular*}
\end{table*}

Moreover, although we have validated the excellent performance of IRM on the implicit advertisement understanding dataset, we remain curious to explore whether IRM can continue to perform better in generalized, real-world video understanding tasks involving more flexible implicit meaning. 
Therefore, we introduce a classical real-world video understanding dataset, Traffic-VQA, for evaluation, where models are required to trace the causes of traffic accidents.
Specifically, we introduce the ``Event Forecasting'' subset of SUTD-Traffic dataset~\cite{sutd}, since it evaluates on predicting implicit outcomes. 
Finally, we collect 2,195 examples for generalization evaluation in a zero-shot manner. As shown in Tab.~\ref{tab: traffic}, IRM also performs SOTA, surpassing VideoChat2 and GPT-4o by 4.18\% and 2.11\%, visualization is shown in Fig.~\ref{fig: vis-traffic}. 
This result proves that IRM can also enhance the performance of answering real-world questions with a certain degree of implicit meaning, such as those related to prediction or planning.
 

Furthermore, we examine whether IRM can effectively reason over more general video questions. In these questions, the distinction between explicit and implicit types is less clear. To this end, we selected several representative tasks from the widely recognized video understanding benchmark MVbench~\cite{videochat2}:

\noindent~1) \textit{``Action Prediction''~(AP)} in the Action Reasoning.

\noindent~2) \textit{``Moving Direction''~(MD)} in the Location Reasoning.

\noindent~3) \textit{``Episodic Reasoning''~(ER)} in the Cognitive Reasoning.

\noindent~4) \textit{``Object Existence''~(OE)} in the Object Reasoning. 

As shown in Tab.~\ref{tab: mvbench}, IRM significantly improves action prediction and motion direction judgment sub-tasks. 
This result aligns with the findings from the evaluation on the SUTD-Traffic~\cite{sutd} dataset, indicating that as the model’s ability to handle implicit questions improves, its predictive and dynamic reasoning capabilities are also significantly enhanced.
Moreover, the overall average accuracy increases by 1.88, demonstrating the strong robustness of our IRM in general videoQA tasks.
However, there is a slight decline in episodic reasoning and object existence inference, indicating that in tasks where visual information is particularly explicit, our model may experience minimal performance degradation.

Overall, on highly general video understanding tasks, our IRM can still balance performance across video questions with varying degrees of implicit meaning, and it continues to provide an overall performance boost.

\section{Conclusion}
\label{sec:conclusion}
We introduce a novel task, Implicit Video Question Answering (I-VQA), which focuses on video reasoning in scenarios where explicit visual evidence is neither directly accessible nor easily locatable. 
Our findings demonstrate that current state-of-the-art Multimodal Large Language Models (MLLMs) struggle to effectively tackle implicit reasoning, underscoring the importance of advancing research in implicit video question answering. 
Additionally, we developed the I-VQA dataset through a robust semi-automated pipeline and proposed the first implicit VideoQA framework, IRM, which integrates dual context clues as auxiliary reasoning paths. 
Experimental results show that IRM outperforms leading MLLMs, such as GPT-4o, fine-tuned Qwen2.5-VL, and VideoChat2 in implicit video question answering. 
Moreover, IRM demonstrates strong performance across multiple VideoQA datasets—PSAV, SUTD-Traffic, and MVBench—underscoring its generalizability to real-world video question answering tasks with varying levels of implicit content.

\section{Statements and Declarations}

All annotations for the I-VQA dataset, including question-answer pairs and context clues annotated by GPT-4o, are now publicly available in the repository: \url{[https://github.com/tychen-SJTU/Implicit-VideoQA}].

The original video sources can be legally obtained from several benchmarks, including Next-GQA~\cite{nextgqa}, E.T. Bench~\cite{etbench}, and REXTIME~\cite{rextime}.


This work was supported in part by the National Natural Science Foundation of China under Grant 62325109, Grant 62561160155, and Grant U21B2013, in part by the Shanghai ‘The Belt and Road’ Young Scholar Exchange under Grant 24510742000, and in part by the National Key R\&D Program of China under Grant 2022ZD0160102.

\newpage
\begin{appendices}
\section{Visualizations}
\label{app_vis}
\begin{figure*}[t]
\begin{center}
\includegraphics[width=0.8\textwidth]{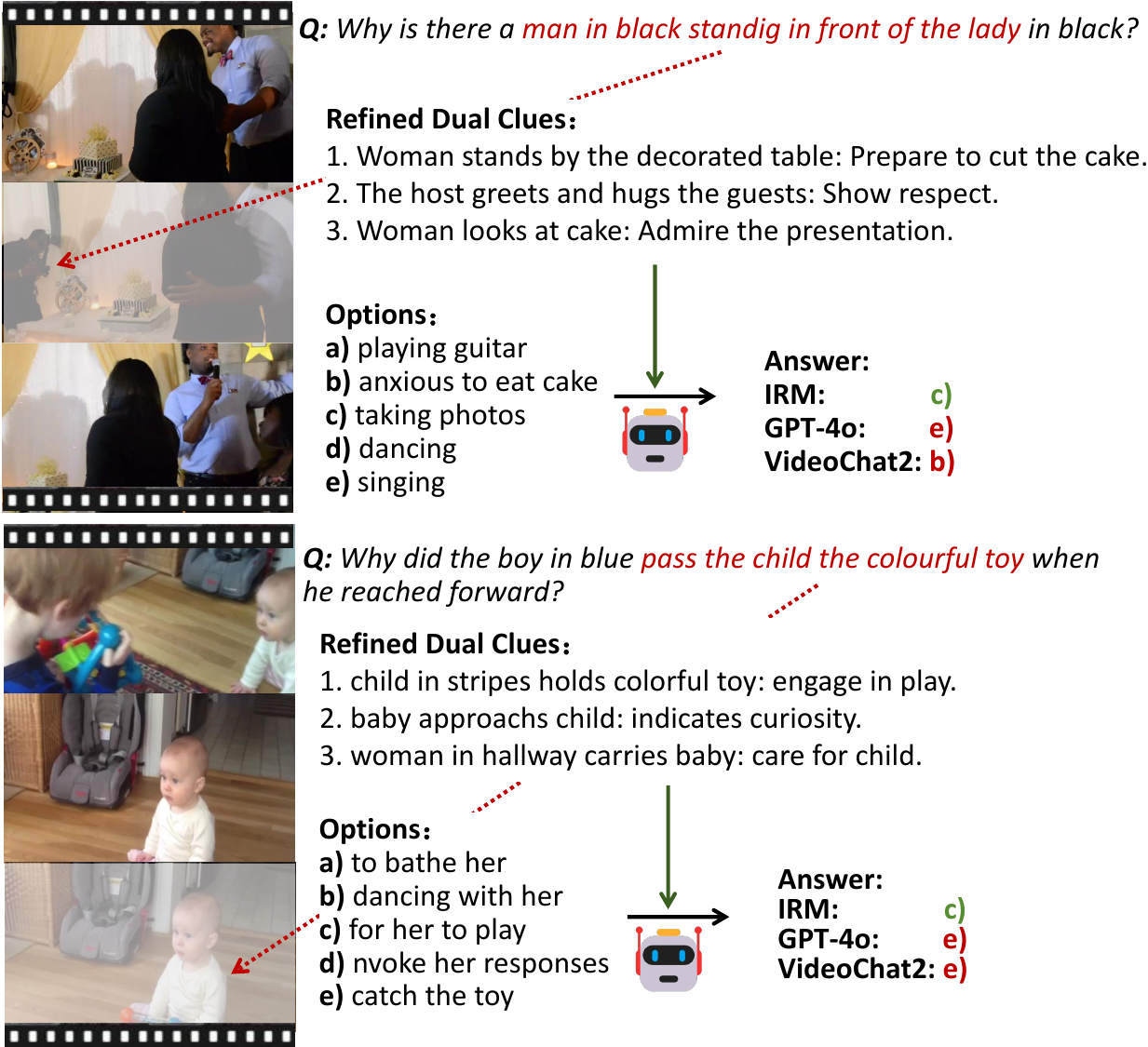}
\end{center}
\vspace{-8pt}
\caption{\textbf{Visualizations of the Multi-choice Setting of the I-VQA Dataset.} The intention of the host's greeting and the action of the woman looking at the cake making ``taking a photo'' to be most reasonable. In the bottom, as the boy plays with a toy and his little sister shows interest, he will likely give it to her for playing next.}
\label{fig: vis-mc_appendix}
\end{figure*}

\begin{figure*}[t]
\begin{center}
\includegraphics[width=0.75\textwidth]{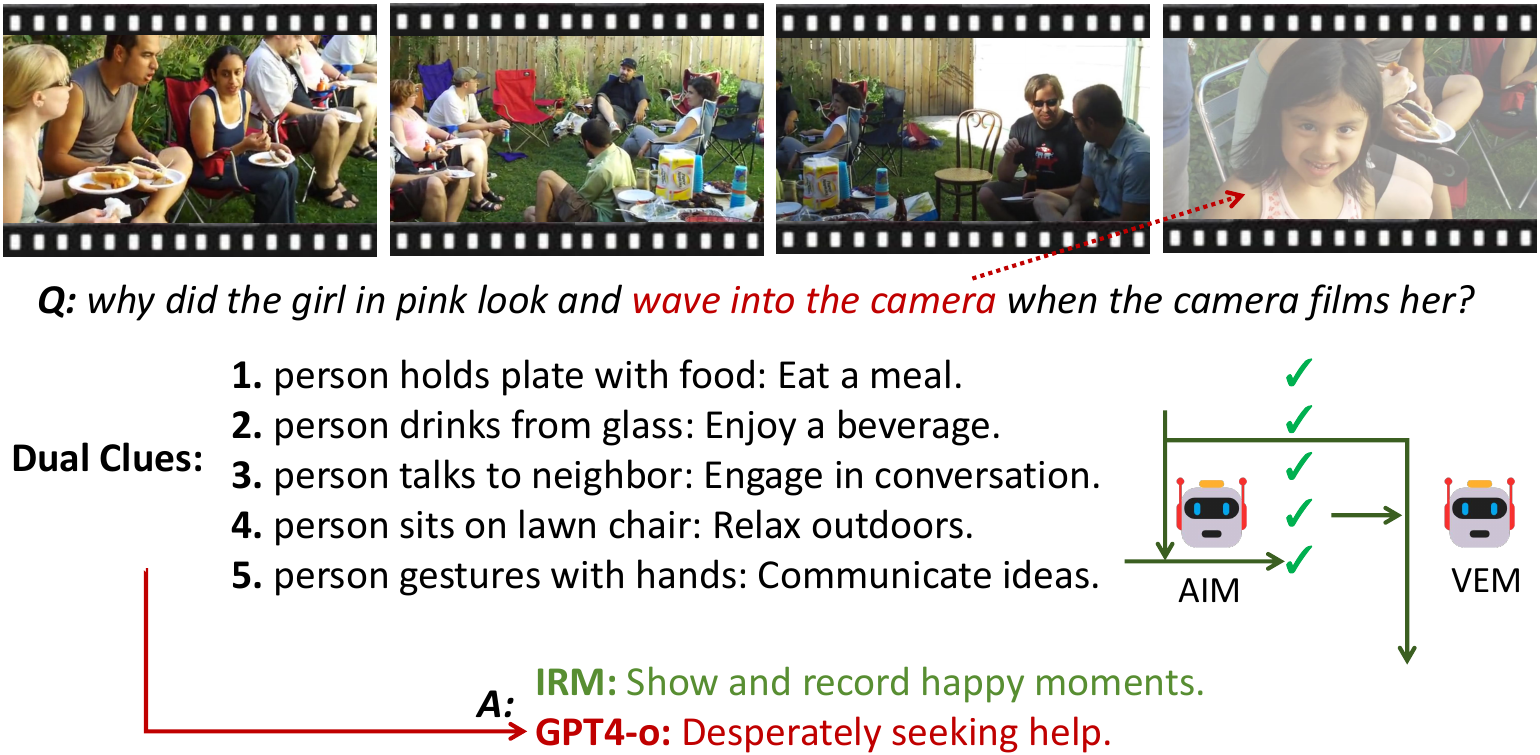}
\end{center}
\vspace{-8pt}
\caption{\textbf{Visualizations of the Open-ended Setting of the I-VQA Dataset.} Our IRM integrates contextual information more effectively while reasoning.}
\label{fig: vis-open_appendix}
\end{figure*}

In the first visualization example of the multi-choice setting of I-VQA in Fig.~\ref{fig: vis-mc_appendix}, the host's gesture of greeting and hugging the guest indicates the end of an activity, not the beginning, thus ruling out options such as singing, dancing, or musical performance. 
Furthermore, the unengaged action between the female guest and the cake suggests that the cake hasn't been cut yet ruling out the intention of eating cake. 
Based on this reasoning, the most plausible scenario is the appearance of the man in black, who is taking a photo of the host and guest.
In the second example, the boy is playing with a toy when his little sister approaches and shows interest in it. Given this interaction, it is highly likely that the boy will give the toy to his sister to play with next.

\begin{figure*}[t]
\begin{center}
\includegraphics[width=0.6\textwidth]{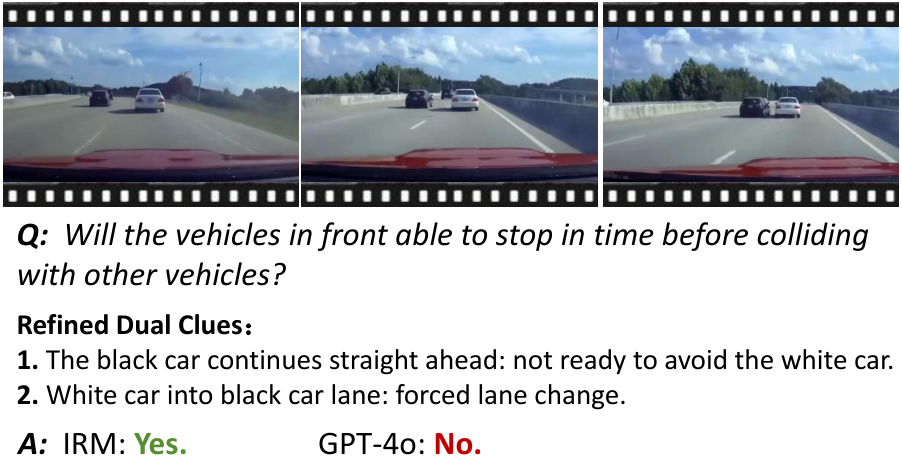}
\end{center}
\vspace{-8pt}
\caption{\textbf{Visualizations of the SUTD-Traffic Dataset.} The routes and intentions of the two vehicles aid in the reasoning of traffic accident prediction.}
\label{fig: vis-traffic_appendix}
\end{figure*}

\begin{figure*}[t]
\begin{center}
\includegraphics[width=0.99\textwidth]{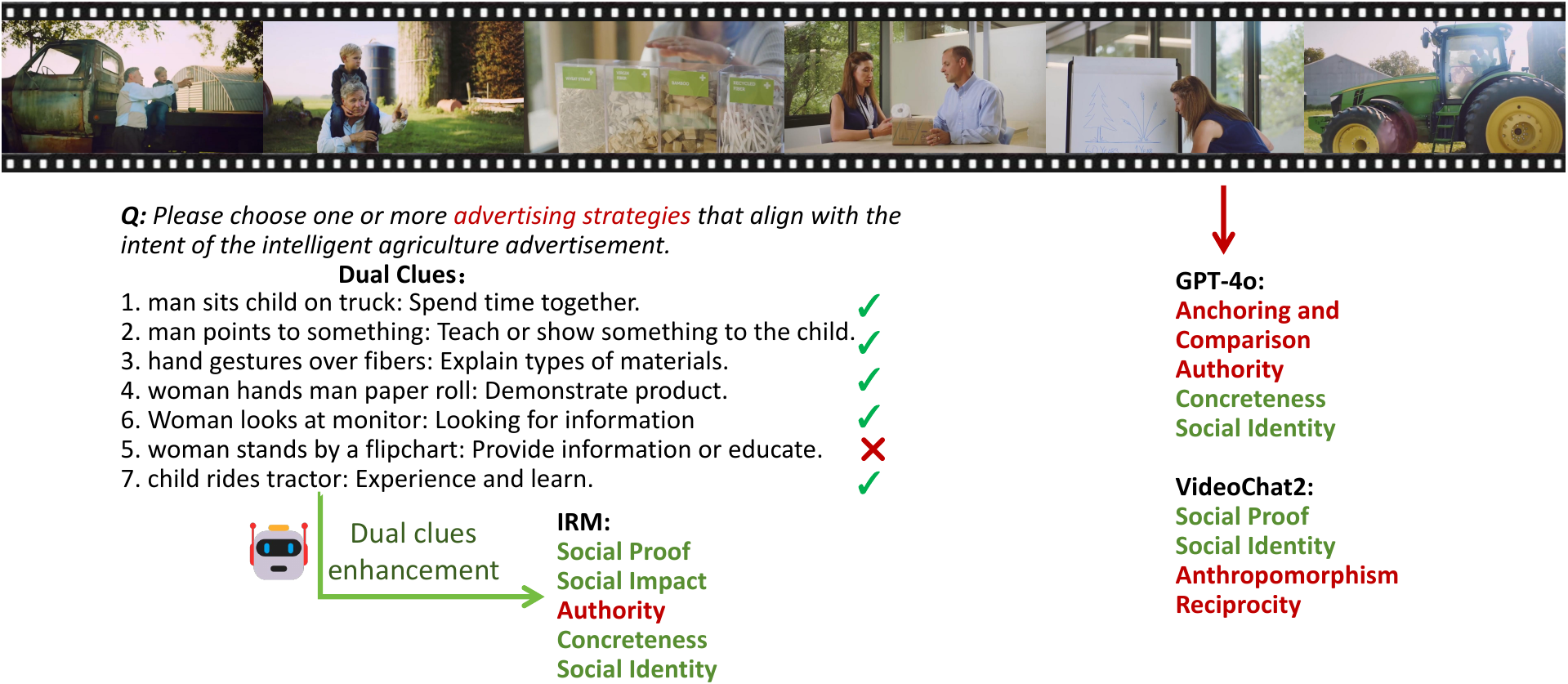}
\end{center}
\vspace{-8pt}
\caption{\textbf{Visualizations of the PSAV Dataset.} Our IRM performs better in both precision and recall when reasoning toward the implicit intent questions.}
\label{fig: vis-psav_appendix}
\end{figure*}

In the visualization example of the open-ended setting of I-VQA in Fig.~\ref{fig: vis-open_appendix}, the context of many people gathering for a meal creates a joyful atmosphere, suggesting that an implicit action, such as waving to the camera, is likely an indication of the intention to capture a happy moment rather than a request for help.
In the SUTD-Traffic dataset experiment shown in Fig.~\ref{fig: vis-traffic_appendix}, IRM successfully identifies the routes and intentions of the two vehicles, accurately predicting the subsequent unseen traffic accidents.

In the advertisement understanding example shown in Fig.~\ref{fig: vis-psav_appendix}, from the perspective of a pair of grandparent and grandchild, the advertisement highlights the social impact of smart agriculture and the public’s recognition of it. 
Our IRM effectively uncovers all the persuasion strategies underlying the event by analyzing the actions and intentions involved. In contrast, other MLLMs fail to grasp the specific dynamics of the grandparent-grandchild interaction or the promotional aspects of smart agriculture.

\section{Inference Speed}
\label{app_speed}
As shown in Tab.~\ref{tab: speed}, the computational overhead introduced by our IRM's context clue refinement and visual clue enhancement is relatively lightweight compared to the overall reasoning process, each sample only takes an additional 0.27 seconds. 

\begin{table*}[t]
  \centering
  \caption{\textbf{Inference Speed.} Our IRM not only achieves SOTA performance but also operates at a relatively fast speed. $*$ indicates with evaluation through API. $\dag$ indicates estimation.}
  \label{tab: speed}
  \begin{tabular*}{\dimexpr 0.85\textwidth}{@{\extracolsep{\fill}}lcc}
    \toprule
    Model & Speed~(seconds/sample) & Model Size \\
    \midrule
    VideoChat2~\cite{videochat2} & 1.68 & 7B \\
    PLLaVA~\cite{pllava} & 1.79 & 7B \\
    VideoLLaVA~\cite{videollava} & 2.02 & 7B \\
    Qwen2.5-VL~\cite{qwen} & 2.27 & 7B \\
    \textbf{IRM} & \textbf{2.51} & 7B \\
    NViLA~\cite{nvila} & 2.55 & 8B \\
    \textbf{IRM with 3 Extra Iterations} & \textbf{3.32} & 7B \\
    Minigpt4-video~\cite{Minigpt4-video} & 8.94 & 7B \\
    DeepSeek-R1~\cite{deepseek} & $~~\text{9.71}^*$ & 671B \\
    OpenAI-o3~\cite{o3} & $~~\text{10.70}^*$ & - \\
    GPT-4o~\cite{gpt4o} & $~~\text{14.11}^*$ & $\text{200B}^\dag$ \\
    \bottomrule
  \end{tabular*}
\end{table*}

\section{Implementation}
\label{app_detail}
\noindent\textbf{Architectures and Training.} 
We fine-tune the visual encoder, the QFormer, and the LLM with question generation loss and relation loss, with initial weights of 1.0 and 2.0. 
The weight of the relation loss decays by 0.05 each epoch, ensuring that the focus of learning gradually shifts to reasoning through clues.

\noindent\textbf{Performances.} 
We report the average results of IRM and all open-source models under three random seeds (2024, 2025, 2026). The default temperature is used for all proprietary models, and the average performance is taken from three experimental runs.

\noindent\textbf{Fine-tuning Details of Comparative Methods.} 
For the fine-tuning of Qwen2.5-VL~\cite{qwen}, we followed the official default implementation by freezing the visual encoder to preserve prior knowledge while keeping the LLM parameters learnable. 
Similarly, for VideoChat2~\cite{videochat2} and PLLaVA~\cite{pllava} fine-tuning, we also followed the official implementations: freezing the visual encoder while training the (Q-Former for VideoChat2) vision-language projection layers, and applying LoRA adapters tuning to the LLM components.

\noindent\textbf{Proprietary Model Version and Input.}  OpenAI-o3 represents the version of ``o3-2025-04-16'', Gemini2.5-Pro represents the version of ``gemini-2.5-pro-preview-05-06'', Claude-3.7-Sonnet represents the version of ``claude-3-7-sonnet-20250219'', GPT-4o represents the version of ``gpt-4o-2024-08-06''. The inputs of Gemini2.5-Pro, GPT-4o, and Claude-3.7-Sonnet are all 8 frames to ensure a fair comparison.

\section{Prompt}
\label{app_prompt}
For the implicit question evaluations on the I-VQA (multi-choice) and the SUTD-Traffic~\cite{sutd} dataset, we design the guiding prompt for IRM and the other comparison models (\textit{\textit{e.g.}}, GPT-4o) as follows:
\begin{lstlisting}
"The question involves implicit visual information, with key visual evidence being invisible, requiring the deduction of the answer based on contextual visual information and provided intention, action clues"
"Context clues are as follows: "
"Clues1: XXX to XXX"
"Clues2: XXX by XXX"
"Clues3: XXX before XXX"
"Based on the clues, select the option that accurately addresses the question."
"Question: XXX"
"(A) XXX ... (E) XXX"
"Only give the best option."
\end{lstlisting}

For the open-ended implicit question evaluations on the I-VQA dataset, we design the guiding prompt for IRM and the other comparison models (\textit{\textit{e.g.}}, GPT-4o) as follows:

\begin{lstlisting}
"The question involves implicit visual information, with key visual evidence being invisible, requiring the deduction of the answer based on contextual visual information and provided intention, action clues"
"Context clues are as follows: "
"Clues1: XXX to XXX"
"Clues2: XXX by XXX"
"Clues3: XXX before XXX"
"Question: XXX"
\end{lstlisting}

Besides, for the implicit question evaluation on the PSAV dataset, we design the guiding prompt for IRM and the other models (\textit{\textit{e.g.}}, GPT-4o) as follows:
\begin{lstlisting}
"Please choose one advertising strategy that aligns with the intent of the advertisement: 
(A) Social Identity
(B) Concreteness
(C) Anchoring and Comparison
(D) Overcoming Reactance
(E) Reciprocity
(F) Foot-in-the-Door
(G) Authority
(H) Social Impact
(I) Anthropomorphism
(J) Scarcity
(K) Social Proof 
(L) Unclear"
"Context clues are as follows: "
"Clues1: XXX to XXX"
"Clues2: XXX by XXX"
"Clues3: XXX before XXX"
"Based on the clues, select the option that accurately addresses the question."
"Only give the best option."
\end{lstlisting}



Additionally, a complete prompt example of IRM under the multi-choice setting of the I-VQA task:

\begin{lstlisting}
[`Human', `The question involves some implicit visual information, with key visual evidence being invisible, requiring the inference of the correct answer based on context. '], 
[`Human', `Context clues are as follows: boy climbs onto apparatus to engage in gymnastics activity, the boy holds onto gymnastic bars to practice balance.'], 
[`Human', `Based on the clues, select the best option that accurately 
addresses the question.'],
[`Human', `Q: Please answer: What does the boy in red and black shorts do as the boy in blue shorts was hanging on the pole? 
(A) walk in front of him
(B) watch the kids
(C) jumps withthem
(D) dropped on table
(E) climb on table'], 
[`Human', `Only give the best option.']
\end{lstlisting}

The prompt templates we used for IRM, GPT-4o, OpenAI-o3 and other models are shown above. For more detailed code implementation, please refer to the open-source code repository \url{https://github.com/tychen-SJTU/Implicit-VideoQA}.

\section{Data Clean Example}
\label{app_clean}
\begin{figure*}[t]
\begin{center}
\includegraphics[width=0.97\textwidth]{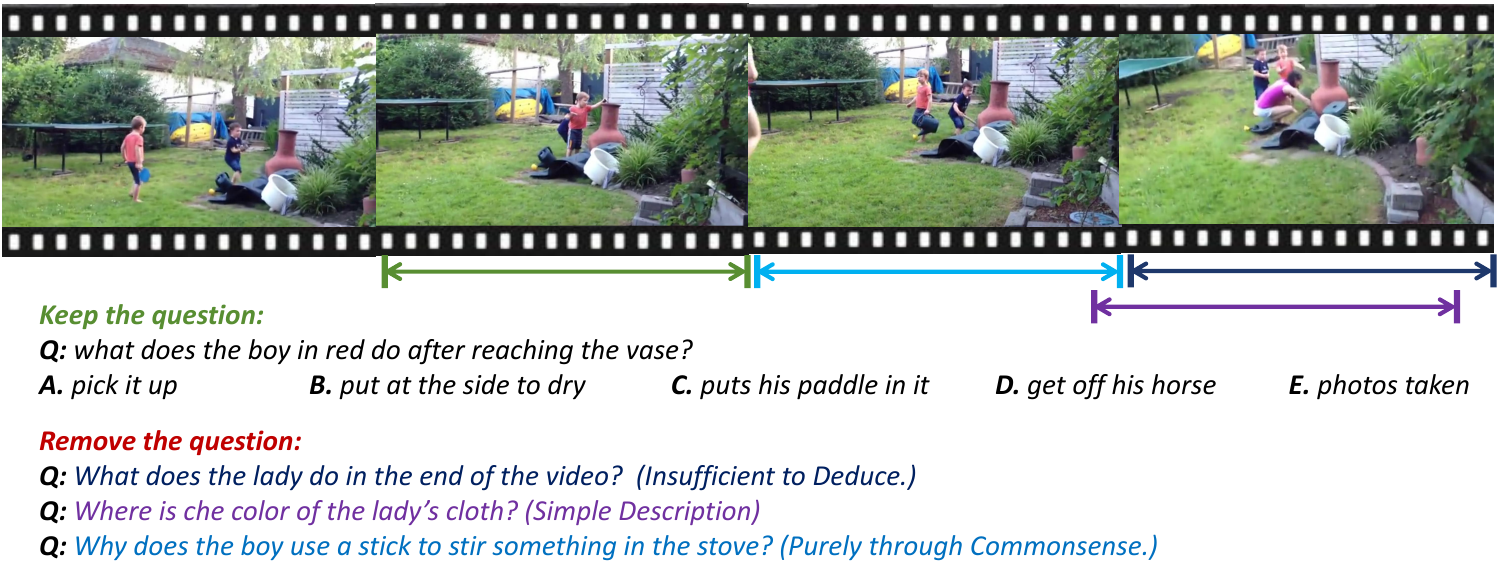}
\end{center}
\vspace{-8pt}
\caption{\textbf{Data Cleaning Example.} We exclude questions that cannot be reasoned using contextual visual information, those that can be adequately answered through commonsense alone, or questions that are simply toward basic descriptions.}
\label{fig: data_lean_appendix}
 \vspace{-7pt}
\end{figure*}

As shown in Fig.~\ref{fig: data_lean_appendix}, we kept the question: \textit{``What does the boy in red do after reaching the vase?''} because, based on the options, we can infer from the context that, earlier, the two boys were holding paddles and walking towards the vase. 
Besides, the two boys are then seen moving the charcoal in the vase with sticks. 
Therefore, the boy approaching the vase is likely trying to add something to it. 
By analyzing the options, it becomes clear that option C is the only logically correct one that can be deduced.

On the other hand, we excluded the question: \textit{``What does the lady do at the end of the video?''} because, based on the previous context, we can only infer that the boy threw the paddle into the vase. However, the actions of their mother/sister afterward cannot be uniquely deduced from the context. She could either educate the boy, attempt to remove the paddle, try to extinguish the fire or do something else entirely. Therefore, the answer is not uniquely inferable.

\section{Examples of DeepSeek-R1}
\label{app_r1}
\textbf{We present the successful reasoning process of DeepSeek-R1 for the question \textit{``What does the boy in red do after reaching the vase?''} shown in Fig.~\ref{fig: data_lean_appendix}:}

\textit{Okay, so I've got this question here that's a bit tricky, but I'll try to work through it step by step. Let me start by understanding the context clues provided.}

\textit{First, there are three main clues:}

\textit{1. A boy in red runs around the garden: Play a game for fun.
2. The boy in blue uses a wooden stick to stir the contents in the vase: Make the fire burn more fiercely.
3. A woman in pink walks toward the boys: Supervise play.}

\textit{Now, the question focuses on what the boy in red does after reaching the vase. So, the boy in red is running around, presumably as part of a game, and when he reaches the vase, he does something specific.}

\textit{Looking at the options:}
\textit{(A) pick it up}
\textit{(B) put at the side to dry}
\textit{(C) puts his wooden paddle in it}
\textit{(D) get off his horse}
\textit{(E) photos taken}

\textit{The boy in red is playing a game, so maybe he's part of a game that involves the vase. The boy in blue is using a wooden stick to stir the vase to make the fire burn more. So, perhaps the vase is part of a fire-related game, like a bonfire or something similar.}

\textit{If the boy in red runs to the vase, what would he do? Option C is about putting a wooden paddle in it. A paddle is like a stick, so maybe he uses it to stir as well, but the boy in blue is already doing that. Alternatively, maybe the boy in red picks it up (option A) to carry it somewhere, but why would he pick it up? Or maybe he puts it aside to dry (B), but that doesn't seem related to a game. Option D is about getting off a horse, which doesn't seem connected. Option E is about taking photos, which also doesn't fit the context of playing a game.}

\textit{So, considering all this, the most plausible answer is option C: puts his wooden paddle in it.
}

\begin{figure*}[t]
\begin{center}
\includegraphics[width=0.99\textwidth]{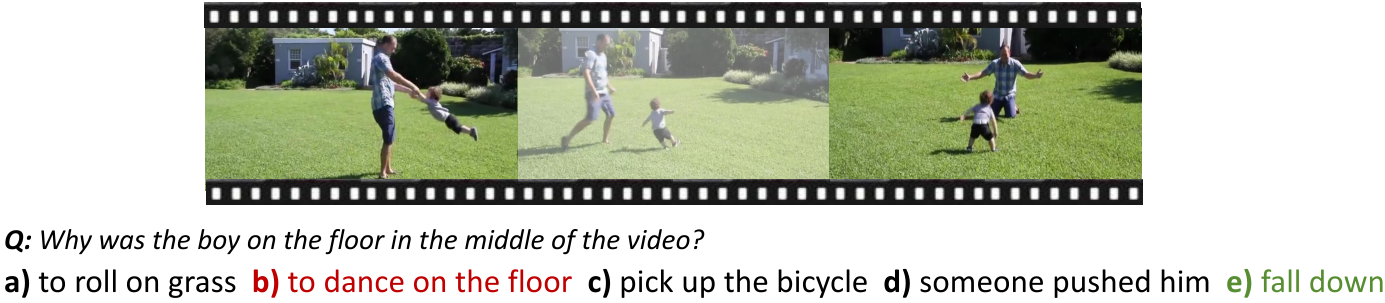}
\end{center}
\vspace{-8pt}
\caption{\textbf{Failure Case Example of DeepSeek-R1.}}
\label{fig: failure_case}
 \vspace{-8pt}
\end{figure*}

Ultimately, through the above analysis, DeepSeek-R1 correctly deduced the boy's intent in approaching the vase, demonstrating implicit reasoning ability supported by contextual clues.
\textbf{Additionally, as shown in Fig.~\ref{fig: failure_case}, we also provide a failure case of DeepSeek-R1's reasoning:}

\textit{Alright, so I have this question about why the boy was on the floor in the middle of the video.}

\textit{First, the context clues provided are:}

\textit{1. A man swings the child: play with the child.
2. The child runs towards the man on the grass: seek affection.
3. The man kneels and opens his arms: invite a hug.
4. The man hugs the child: show affection.}

\textit{So, the scenario seems to involve interaction between a man and a child, specifically the boy. The actions described are playful and affectionate.}

\textit{The question is why the boy was on the floor. The options are about different reasons. Let me think about each option in relation to the context clues.}

\textit{Option A: Wanted to roll on grass. Well, the context mentions the man swinging the child, which is a play activity. Maybe they were outside, but the clues don't specify grass. So, I'm not sure about this one.}

\textit{Option B: To dance on the floor. Hmm, if the man is playing with the child, maybe they were dancing together. The man kneeling and hugging could be part of dancing moves. That makes sense because dancing often involves being on the floor or moving around.}

\textit{Option C: Pick up the bicycle. The context doesn't mention anything about a bicycle. So, probably not this one.}

\textit{Option D: Someone pushed him. The clues don't indicate any negative action like pushing. It's more about affectionate play.}

\textbf{\textit{Option E: Fall down. While playing, a child may not fall with the care of his father.}}

\textit{Now, focusing on option B: dancing on the floor. The clues show that the man is actively engaging with the child, which could include dancing. The actions described, like swinging, running towards, kneeling, and hugging, are all part of playful interaction, which could very well include dancing. \textbf{So, it makes sense that the boy was on the floor because they were dancing together.}}

DeepSeek-R1~\cite{deepseek} incorrectly assumes that the presence of a father’s care would prevent falling, instead choosing the option of dancing on the floor, which seems superficially related to the ``floor''.

\section{Score Coherence Evaluation}
\label{app_score}
To ensure that the GPT-Score we use for open-ended evaluation is reasonable and consistent with traditional VideoQA, we calculated the Pearson correlation coefficient~\cite{pearson} between the average scores given by five annotators and the scores assigned by GPT-3.5. 
We found that the coefficient reached as high as 0.93 (with Pearson correlation values ranging from -1 to 1, where values closer to 1 indicate stronger correlation), demonstrating that this metric is also applicable to our proposed implicit-VQA.

\section{Datasets}
\label{app_dataset}
\subsection{Details of the GroundedVQA Datasets}
Next-GQA comprises 10,531 annotated evidence temporal segments, corresponding to 8,911 QA pairs and 1,557 videos from NextQA~\cite{nextqa} which consists of a subset focused on complex temporal and causal reasoning. 
E.T. Bench includes 10,000 grounded QA samples derived from 7,002 self-collected videos, requiring models to demonstrate event-level comprehension and time-sensitive reasoning. 
REXTIME consists of 12,000 QA samples with annotated answer temporal locations, sourced from ActivityNet~\cite{acititynet} and QV-Highlights~\cite{qv-highlight}. It is specifically designed to evaluate the ability to perform temporal reasoning.

\subsection{Details of the PSAV Dataset}
The  ``Persuasion Strategies in Advertisement Videos'' dataset~\cite{adsdataset} (PSAV) contributes an important task on the advertisement story understanding task, namely persuasion strategy identification. 
For this task, the authors collect 1,002 video advertisements from popular brands available on the web publicly and utilize the persuasion strategy labels defined by~\cite{Persuasion}. 
The 12 strategies target persuasion strategy set are: Social Identity, Concreteness, Anchoring and Comparison, Overcoming Reactance, Reciprocity, Foot-in-the-Door, Authority, Social Impact, Anthropomorphism, Scarcity, Social Proof, and Unclear.

\subsection{Details of the SUTD-Traffic Datasets}
The SUTD-TrafficQA~\cite{sutd} dataset is a large-scale collection of 10,080 videos from diverse real-world traffic scenarios, sourced both online (YouTube, LiveLeak, Twitter, Bili-bili) and offline (handheld cameras and car-mounted recorders) to ensure global diversity. 
It contains 62,535 human-generated question-answer pairs across six complex reasoning tasks: basic event understanding, event forecasting, reverse reasoning, counterfactual inference, introspection, and attribution. These tasks focus on causal inference, event understanding, and multi-level logical reasoning.

\section{Limitations and Future Work}
\label{app_future}

1. We have demonstrated that the thinking patterns related to invisible or deep-level information in implicit questions are beneficial for event/action prediction tasks, as they similarly involve predicting the unknown. 
In the future, it can also be further explored whether including prediction tasks in the training phase could similarly enhance implicit reasoning abilities.


2. We validated the generalization of our model on the implicit video question answering we proposed, as well as on a broader range of video tasks (including advertisement classification, video event prediction, and explicit video question answering). However, we did not validate it on more other tasks, as they are less related, leaving that exploration for future research.

3. Although our model relies on GPT-4o-generated candidate clues during training, our IRM operates completely independently during inference. Additionally, we have open-sourced the contextual clues to facilitate future research.

4. We only conducted experiments on scaling up the training data and inference time. Due to limited computational resources, the investigation of scaling up the model itself is left for future research.

\end{appendices}
\newpage

\bibliography{sn-bibliography}%
\end{document}